\pdfoutput=1

\documentclass[11pt]{article}

\usepackage[final]{acl}

\usepackage{times}
\usepackage{latexsym}

\usepackage{kotex}
\usepackage{listings}
\usepackage{xcolor}
\usepackage{makecell}

\usepackage[utf8]{inputenc}
\usepackage{amsmath}
\usepackage{amsfonts}
\usepackage{algorithm}            
\usepackage{algorithmicx}         
\usepackage{algpseudocode}
\usepackage{url}

\usepackage{xcolor}
\usepackage{soul}
\definecolor{highlightyellow}{HTML}{FFFC9E}
\sethlcolor{highlightyellow}

\usepackage{tabularx}
\usepackage{array}
\usepackage{graphicx}

\usepackage{caption}

\usepackage[T1]{fontenc}

\usepackage[utf8]{inputenc}

\usepackage{inconsolata}

\usepackage{graphicx}
\usepackage{booktabs}
\usepackage{colortbl}
\usepackage{hyperref}
\usepackage{enumitem}

\usepackage[verbose=silent]{microtype}
\usepackage{microtype}

\newcommand{\datasetName}{\textsc{KoBLEX}}
\newcommand{\evalName}{\textsc{LF-Eval}}
\newcommand{\methodName}{\textsc{ParSeR}}

\title{\datasetName: Open Legal Question Answering with Multi-hop Reasoning}

\author{
    Jihyung Lee
    \thanks{Equal    Contribution.}$^1$, 
    Daehui Kim
    \footnotemark[1]$^{1,3}$, 
    Seonjeong Hwang$^1$, 
    Hyounghun Kim$^{1,2}$, 
    Gary Geunbae Lee$^{1,2}$ \\
    $^1$Graduate School of Artificial Intelligence, POSTECH, Republic of Korea\\
    $^2$Department of Computer Science and Engineering, POSTECH, Republic of Korea\\
     $^3$AI Future Lab, KT, Republic of Korea\\
    \texttt{\{jihyung.lee,  
    andrea0119, 
    seonjeongh,
    h.kim, 
    gblee\}@postech.ac.kr} \\
}

\begin{document}

\maketitle
\begin{abstract}
Large Language Models (LLM) have achieved remarkable performances in general domains and are now extending into the expert domain of law. Several benchmarks have been proposed to evaluate LLMs' legal capabilities.
However, these benchmarks fail to evaluate open-ended and provision-grounded Question Answering (QA).
To address this, we introduce a \textbf{Ko}rean \textbf{B}enchmark for \textbf{L}egal \textbf{EX}plainable QA (\textbf{\datasetName}), designed to evaluate provision-grounded, multi-hop legal reasoning.
\datasetName\ includes 226 scenario-based QA instances and their supporting provisions, created using a hybrid LLM–human expert pipeline.
We also propose a method called \textbf{Par}ametric provision-guided \textbf{Se}lection \textbf{R}etrieval (\textbf{\methodName}), which uses LLM-generated parametric provisions to guide legally grounded and reliable answers. \methodName{} facilitates multi-hop reasoning on complex legal questions by generating parametric provisions and employing a three-stage sequential retrieval process.
Furthermore, to better evaluate the legal fidelity of the generated answers, we propose \textbf{L}egal \textbf{F}idelity \textbf{Eval}uation (\textbf{\evalName}). \evalName{} is an automatic metric that jointly considers the question, answer, and supporting provisions and shows a high correlation with human judgments.
Experimental results show that \methodName{} consistently outperforms strong baselines, achieving the best results across multiple LLMs. 
Notably, compared to standard retrieval with GPT-4o, \methodName{} achieves 37.91 higher F-1 and 30.81 higher \evalName.
Further analyses reveal that \methodName{} efficiently delivers consistent performance across reasoning depths, with ablations confirming the effectiveness of \methodName. \footnote{The code and dataset are available at \url{https://github.com/daehuikim/KoBLEX}} 
\end{abstract}
\section{Introduction}\vspace{-7pt}

\begin{figure}
    \centering
    \includegraphics[width=\linewidth]{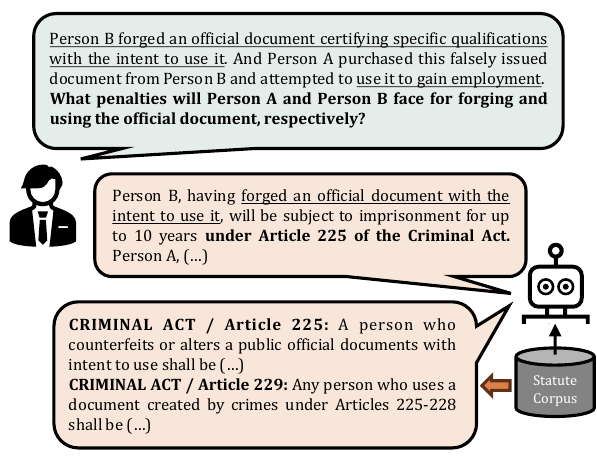}
    \caption{\textbf{Overview of \datasetName\ structure and task design.}  
Given a complex legal question, the system is required to reason over multiple statutory provisions.}
    \label{fig:overview_figure}
    \vspace{-7pt}
\end{figure}

Large Language Models (LLM) have demonstrated strong performance across a wide range of tasks \cite{zhao2025surveylargelanguagemodels}, leading to the development of diverse benchmarks across general domains \cite{hendryckstest2021,cobbe2021gsm8k,zhou2023instruction,NEURIPS2023_91f18a12,rein2024gpqa}.
As LLMs increasingly demonstrate expert-level capabilities, interest in their application to the legal domain has grown, leading to the development of several legal benchmarks \cite{peng2023study,sun2023short,fei-etal-2024-lawbench,guha2023legalbench,li2024lexeval}.
Although recent legal benchmarks offer a diverse set of tasks to assess LLMs’ capabilities in the legal domain, they are not well-suited for evaluating open-ended and provision-grounded legal question answering (QA) \citep{son2024kmmlu, kimyeeun-etal-2024-developing}.
In practice, users often pose complex legal questions and expect answers grounded in legal provisions, as illustrated in Figure \ref{fig:overview_figure}.
Such grounding is critically essential in the legal domain, where hallucinated or inaccurate information can easily lead to serious situation \cite{engstrom2021legal,romoser2023no,dahl2024large}.
However, generating responses grounded in legal provisions is challenging because it requires not only identifying relevant provisions but also interpreting them with sufficient expert knowledge.

In light of these limitations, there is a clear need for a comprehensive evaluation of open-ended, provision-grounded legal QA.
To this end, we present a \textbf{Ko}rean \textbf{B}enchmark for \textbf{L}egal \textbf{EX}plainable open-ended QA (\datasetName), designed to evaluate multi-hop legal reasoning capabilities.  
\datasetName\ comprises 226 multi-hop questions, answers, and their supporting statutory provisions, curated through a hybrid pipeline that combines LLM-based generation with expert revision and evaluation. While strict filtering pipeline leads to a limited sample size, experimental results show that \datasetName{} serves as an effective benchmark for distinguishing reasoning capabilities across diverse methods.
Moreover, unlike traditional legal benchmarks that rely on simple matching tasks or multiple-choice questions, \datasetName{} evaluates methods’ ability to generate free-form answers grounded in legal provisions. 
To promote accessibility and multilingual research, all instances are provided in both Korean and English. 
As illustrated in Figure~\ref{fig:overview_figure}, \datasetName{} facilitates the evaluation of provision retrieval accuracy and multi-hop legal reasoning based on generated free-form answers.

Given the knowledge-intensive nature of provision-grounded legal QA, accurate retrieval is critical for generating factual answers. In light of this, we introduce \textbf{Par}ametric provision-guided \textbf{Se}lection \textbf{R}etrieval (\methodName).
\methodName{} first generates \textit{parametric provisions}, provisions constructed using the LLM’s parametric knowledge to emulate the structure and language of real statutes. These serve as query scaffolds to improve the retrieval of relevant legal provisions. 
\methodName{} then identifies the most relevant provision through a three-stage Retrieve–Rerank–Selection retrieval to answer the open-ended complex legal question.

To reliably assess whether the generated answers are faithful to the question,  we also propose \textbf{L}egal \textbf{F}idelity \textbf{Eval}uation (\evalName).
\evalName{} is built on the G-Eval \cite{liu-etal-2023-g}, using instances from \datasetName{}  to assess legal fidelity.
\evalName{} shows robust performance, achieving a Pearson correlation of 84.90 with human judgments.

Experimental results demonstrate that \methodName{} outperforms strong retrieval-augmented reasoning baselines by consistently delivering performance gains across multiple LLMs and diverse evaluation metrics, including \evalName{}. Notably, with GPT-4o, \methodName{} improves provision retrieval accuracy over one-time retrieval by +37.91 F-1 and +19.91 EM and enhances answer quality by +19.39 token-level F-1 and +30.81 \evalName{}, demonstrating its effectiveness.
\methodName{} consistently outperforms baselines across different reasoning depths.
Further ablation analysis reveals that each component of \methodName{} contributes to performance gains.
In addition, \methodName{} demonstrates greater efficiency compared to other baselines, achieving consistently strong results while generating much fewer tokens.

\noindent Our contributions are summarized as follows:
\begin{itemize}[leftmargin=10pt, itemsep=1pt, topsep=0pt]

    \item We introduce \textbf{\datasetName}, a bilingual Korean-English benchmark of 226 provision-grounded, multi-hop legal QA instances curated via LLM and human validation pipeline.

    \item We introduce \textbf{\methodName}, which combines LLM-generated parametric provisions with a three-stage retrieval pipeline, significantly outperforming existing retrieval-augmented reasoning baselines across multiple LLMs.
    
    \item We propose \textbf{\evalName}, a legal fidelity evaluation metric that excels at assessing the legal accuracy and provision alignment of generated responses.

    \item Experiments and analyses demonstrate the effectiveness and efficiency of \methodName{} across models, metrics, and reasoning depths.
    
\end{itemize}

\begin{figure*}
\begin{center}
   \includegraphics[width=\linewidth]{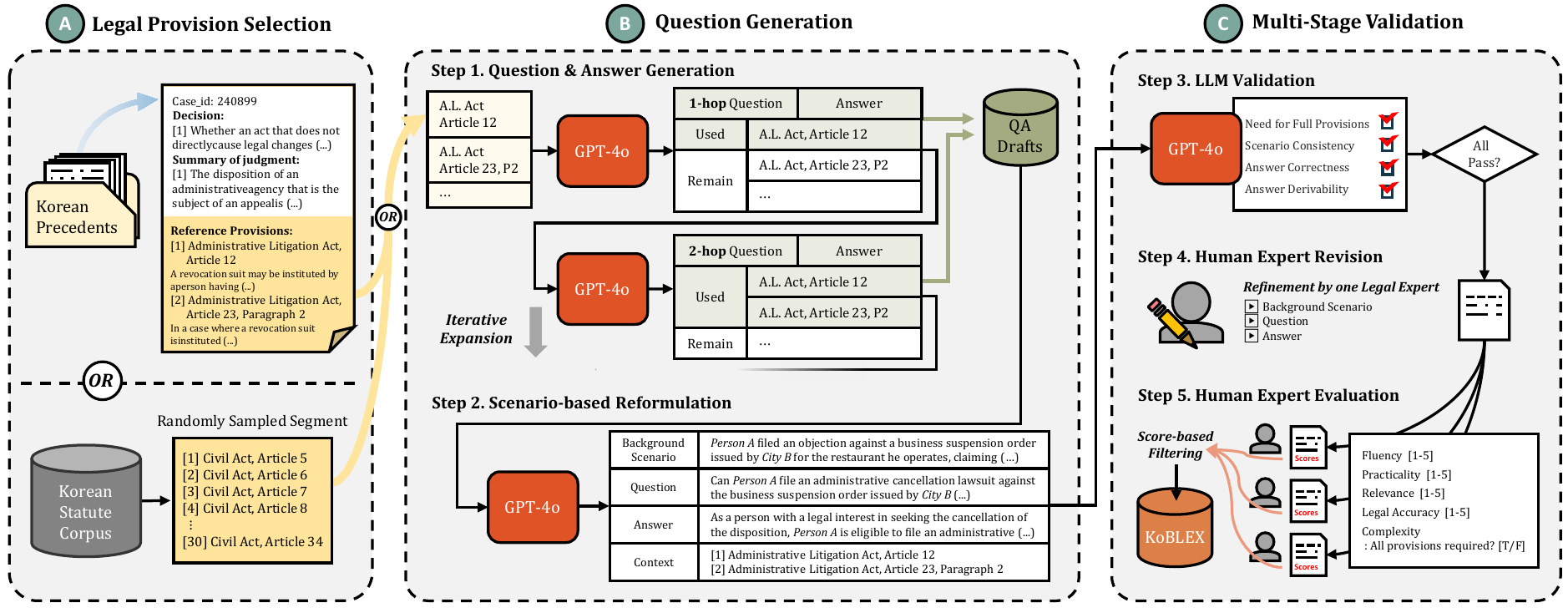}
\end{center}
\vspace{-7pt}
   \caption{\textbf{Generation and validation pipeline for \datasetName.} The pipeline consists of three stages: (A) context construction from either randomly sampled segment of statute corpus or reference provisions in precedents, (B) question–answer generation using GPT-4o based on the selected legal context, and (C) multi-stage validation. An initial LLM-based evaluation filters out incomplete or unsupported pairs based on predefined criteria. Subsequently, one legal expert manually revised each validated pair and three legal experts rated them along five dimensions: fluency, practicality, relevance, legal accuracy, and complexity.}
\label{fig:qg_pipeline}
\vspace{-7pt}
\end{figure*}

\section{Related Works}
\paragraph{Legal Benchmarks.}
A growing body of work has developed benchmarks for evaluating LLMs in legal contexts, spanning a diverse range of jurisdictions, languages, and task formats \citep{kim2016coliee, chalkidis-etal-2019-neural, zhong2020jec, tuggener-etal-2020-ledgar, chalkidis-etal-2022-lexglue, kapoor-etal-2022-hldc, chalkidis2023lexfiles, zhang-etal-2023-fedlegal,li2024legalagentbench}.  
LegalBench \citep{guha2023legalbench} proposes 162 few-shot tasks in English across six categories of legal reasoning, offering one of the most comprehensive benchmark suites to date.  
LawBench \citep{fei-etal-2024-lawbench} adapts this paradigm to Chinese law, comprising 20 tasks that assess capabilities in judgment prediction, statutory interpretation, and legal knowledge retrieval. 

In the Korean legal domain, several benchmarks have been introduced to evaluate domain-specific language understanding.  
LBOX OPEN \citep{hwang2022multi} presents a large-scale multi-task benchmark constructed from Korean court decisions, encompassing tasks such as classification and judgment prediction.  
KMMLU \citep{son2024kmmlu} includes Korean legal QA as a multiple-choice category within a broader zero-shot evaluation suite. 
KBL \citep{kimyeeun-etal-2024-developing} offers a diverse set of multiple-choice legal tasks, including bar exam and scenario-based questions.

However, all of these benchmarks share key limitations: they rely on multiple-choice or binary formats, lack explicit links to relevant statutory provisions, and do not support open-ended QA. 
As a result, they are not suitable for evaluating models’ ability to produce explainable and factually grounded answers in complex legal settings.
These limitations motivate the development of \datasetName, a Korean benchmark for open-ended legal QA that requires provision-grounded, open-ended and multi-hop reasoning over statutory provisions.

\paragraph{Retrieval Augmented Reasoning.} 
Standard Prompting (SP) was proposed to show that LLMs can leverage their parametric knowledge through in-context learning \cite{NEURIPS2020_1457c0d6}.
Building on this, Chain of Thought (CoT) enables step-by-step reasoning by explicitly activating the model’s parametric knowledge \cite{NEURIPS2022_9d560961}.
However, LLMs often fail to align with knowledge-intensive factual information, leading to hallucination \cite{10.1145/3703155}.
To address this limitation, researchers have increasingly explored Retrieval-Augmented Generation (RAG), which enhances response accuracy by incorporating relevant external knowledge \cite{NEURIPS2020_6b493230, gao2024retrievalaugmentedgenerationlargelanguage, cho-lee-2025-k}.
While RAG leverages external knowledge, it often fails to integrate external knowledge with the model’s parametric knowledge, limiting its performance on complex reasoning tasks.
To overcome this, recent work has introduced Retrieval-Augmented Reasoning (RARE), which aims to combine retrieval with multi-step reasoning capabilities. 
Self-Ask~\cite{press-etal-2023-measuring} introduces iterative RARE by generating intermediate questions and querying external knowledge to derive the final answer.
IRCoT~\cite{trivedi-etal-2023-interleaving} interleaves CoT traces with retrieval, using the interleaved generations to incorporate external knowledge into the reasoning process.
FLARE~\cite{jiang-etal-2023-active} improves upon IRCoT by using the model’s token-level confidence to adaptively retrieve contexts.
ProbTree~\cite{cao-etal-2023-probabilistic} decomposes complex questions into a tree of sub-queries, solves each node using diverse strategies, and aggregates the results based on token-level confidence to produce the final answer.
BeamAggr~\cite{chu-etal-2024-beamaggr} enhances ProbTree by performing multi-source reasoning to generate answer candidates at leaf nodes, followed by beam combination and probabilistic answer aggregation.
Despite advances in complex multi-hop reasoning, these approaches remain underexplored in knowledge-intensive domains like law.
This underscores the need for methods that effectively handle complex multi-hop reasoning in knowledge-intensive domains.
To bridge this gap, our research propose \methodName{}, a novel framework that integrates the parametric knowledge of LLMs with a 3-stage retrieval pipeline to support effective retrieval-augmented reasoning.
\begin{figure*}[ht]
\centering
\small
\renewcommand{\arraystretch}{1.4}
\setlength{\tabcolsep}{5pt}
\begin{tabular}{|p{3.5cm}|p{11.8cm}|}
\hline
\textbf{Background Scenario} &
Person B forged an official document certifying specific qualifications with the intent to use it. Person A purchased this falsely issued document from Person B and attempted to use it to gain employment. \\
\hline
\textbf{Question} &
What penalties will Person A and Person B face for forging and using the official document, respectively? \\
\hline
\textbf{Answer} &
Person B, having forged an official document with the intent to use it, will be subject to imprisonment for up to 10 years under Article 225 of the Criminal Act. Person A, having used a document created in violation of Article 225, will be subject to imprisonment for up to 10 years under Article 229 of the Criminal Act. \\
\hline
\textbf{Reference Provision [1]} &
\textbf{CRIMINAL ACT / Article 225}: A person \hl{who counterfeits or alters a public official document} with intent to use shall be punished by imprisonment with labor for not more than ten years. \\
\hline
\textbf{Reference Provision [2]} &
\textbf{CRIMINAL ACT / Article 229}: Any person who \hl{uses a document created by crimes under} %
\hl{Articles 225\textrm{--}228} shall be punished by the penalty prescribed for such crimes. \\
\hline
\end{tabular}
\caption{\textbf{Example QA instance from the \datasetName\ (translated from Korean).}  
This multi-hop question requires interpreting multiple statutes. \hl{Yellow texts} highlight key legal information essential for deriving the correct answer.}
\label{fig:qa_example}
\vspace{-7pt}
\end{figure*}

\section{\datasetName} \vspace{-7pt}
In this section, we describe the construction process of \datasetName{}.
We first generate initial drafts using an LLM, then filter and revise them through a multi-stage validation process involving both LLM-based and human expert review (Figure \ref{fig:qg_pipeline}).

\subsection{Legal Provision Selection} \vspace{-4pt}
Korean statutes are systematically structured at the article and paragraph level, and adjacent provisions typically exhibit strong topic continuity by addressing the similar legal concept or regulatory subject \citep{moleg2024drafting}.  
Accordingly, sampling a continuous segment is a reasonable heuristic for selecting relevant legal content.  
However, this approach may limit the diversity of questions spanning multiple legal sources.

To complement this approach, we also extract statutes cited in real-world court decisions. Specifically, we utilize the \textit{reference provisions} field in Korean precedents to identify statutory clauses that are actually invoked in judicial reasoning. Data sources are described in Appendix~\ref{sec:appendix_license}.

\subsection{Question Generation} \vspace{-4pt}
\paragraph{Question \& Answer Generation.}
\label{sec:qa_construction}
Based on the statutory provisions selected in the previous step, we use GPT-4o \cite{openai2024gpt4technicalreport} to generate initial drafts of question-answer pairs.

As illustrated in Step 1 of Figure \ref{fig:qg_pipeline}, we adapt an incremental generation strategy. We first prompt the model to generate single-hop questions, each of which can be answered using only one provision. Then, based on these single-hop questions, we incrementally expand them into two-hop and three-hop versions by introducing additional provisions and guiding the model to integrate them logically into the reasoning process.

\paragraph{Scenario-based Reformulation.} 
\label{sec:scenario_reformulation}
To better reflect realistic legal situations, we reformulate each question–answer pair into a fact-based legal scenario using GPT-4o. This transformation encourages more natural multi-hop reasoning and enhances the dataset's alignment with practical legal contexts.
Accordingly, each QA draft is rewritten as a scenario-based item structured as realistic legal facts and anonymized parties (e.g., Person A). 

After this process, each QA instance is structured into four components:  
Background Scenario $B$, Question $Q$, Answer $A$, Context $C$,  
where $C = \{p_1, p_2, \dots, p_n\}$ denotes the set of reference statutory provisions used to support the reasoning. A representative QA instance is shown in Figure \ref{fig:qa_example}.

\subsection{Multi-Stage Validation}
\paragraph{LLM Validation.}
\label{sec:llm_validation}
Inspired by prior work that uses LLMs as evaluators for quality control \citep{bedi2024quest}, we employ GPT-4o to filter incorrect QA samples before the human validation step.  
In this step, as illustrated in Step 3 of Figure \ref{fig:qg_pipeline}, the LLM automatically evaluates whether each question $Q$ requires all $C$, and the consistency of $B$, $Q$, and $A$. We term each step as \textit{Partial Check} and \textit{Full Check}.
\textit{Partial Check} is designed to ensure that $Q$ genuinely requires the full set of $C$ for resolution.  
Given $C = \{p_1, p_2, \dots, p_n\}$, we evaluate whether any non-empty subset of the powerset of $C$ is sufficient to answer $Q$.  
If any such subset yields a correct answer without referencing the remaining provisions, we consider the instance to lack true multi-hop characteristics and exclude it.
\textit{Full Check} involves a comprehensive validation of the triplet $(B, Q, A)$ through an inclusive single evaluation prompt, covering the following three aspects:  
(1) \textit{Scenario consistency}, which assesses whether $(B, Q)$ is logically and legally coherent with $C$;  
(2) \textit{Answer correctness}, which determines whether $A$ aligns with the statutory interpretation of $C$; and  
(3) \textit{Answer derivability}, which checks whether $A$ can be fully inferred from $C$ without requiring any unstated assumptions.  
The prompt templates used for both checks are provided in Appendix~\ref{sec:appendix_prompt_templates}.
 
\paragraph{Human Expert Revision.} To ensure legal correctness, clarity, and linguistic fluency, QA instances are further revised and verified by Korean law school graduates and students.  
Each QA instance is classified as either \textit{Pass}, \textit{Revise}, or \textit{Hold}.  
Instances marked as \textit{Revise}—such as ambiguous legal actors in the scenario or legally incorrect answers—are corrected accordingly. In the case of \textit{Hold}, most instances are excluded from the final dataset unless the issue can be resolved by appending additional statutory provisions. Details about human expert revision are in Appendix~\ref{sec:appendix_annotator_guidelines}. 

\paragraph{Human Expert Evaluation.} \label{sec:human_eval} 
Following the revision, each QA instance is evaluated by three different legal experts using five criteria: \textit{Fluency}, \textit{Practicality}, \textit{Relevance}, \textit{Legal Accuracy}, and \textit{Complexity}.  
Fluency, practicality, relevance, and legal accuracy are rated on a 5-point Likert scale, while complexity is a binary label (True/False), with True indicating that the question requires utilizing the full set of $C$.
Descriptions of each criterion and the evaluation guidelines are provided in Appendix~\ref{sec:appendix_annotator_guidelines}.

Instances that show no critical issues during revision and received sufficiently high evaluation scores across these criteria are curated for inclusion in the final dataset.  
Details of the filtering process are provided in Appendix~\ref{sec:dataset_filtering_criteria}.

\subsection{Data statistics}

\begin{table}[t]
\resizebox{\columnwidth}{!}{%
\begin{tabular}{rr|rrr}
\hline
 &  & \multicolumn{3}{c}{\textbf{\# of words}} \\ \cline{3-5} 
 & \textbf{\# Examples} & \multicolumn{1}{r|}{\textbf{\begin{tabular}[c]{@{}r@{}}Background\\ scenario\end{tabular}}} & \multicolumn{1}{r|}{\textbf{Question}} & \textbf{Answer} \\ \hline
1-hop & 55 & \multicolumn{1}{r|}{33.96} & \multicolumn{1}{r|}{13.35} & 11.67 \\ \hline
2-hop & 125 & \multicolumn{1}{r|}{37.45} & \multicolumn{1}{r|}{18.78} & 29.56 \\ \hline
3-hop & 46 & \multicolumn{1}{r|}{46.2} & \multicolumn{1}{r|}{27.13} & 53.11 \\ \hline
\textbf{Total} & 226 & \multicolumn{1}{r|}{38.38} & \multicolumn{1}{r|}{19.16} & 30.0 \\ \hline

\end{tabular}
}
\caption{\textbf{Statistics on \datasetName{}.} The average numbers of words on each QA component are categorized by reasoning depth (1-hop, 2-hop, and 3-hop).}
\label{tab:KoBLEX-Eval_statistics}
\vspace{-7pt}
\end{table}

\datasetName{} comprises 226 high-quality QA instances.   
Table \ref{tab:KoBLEX-Eval_statistics} summarizes the number of examples and the average word count for each QA component grouped by reasoning depth.  
As reasoning depth increases, the average word count consistently rises across all components, clearly reflecting the greater linguistic and logical complexity involved in multi-hop legal reasoning.

The final benchmark spans 83 distinct Korean statutes, including major codes such as the \textit{Civil Act}, \textit{Criminal Act}, and \textit{Criminal Procedure Act}.  
A complete list of statutes and their distribution is provided in Appendix \ref{sec:appendix_distribution_statutes}.  
Among the 226 QA instances, 153 are generated using reference provisions extracted from Korean court decisions, while the remaining 73 are constructed from randomly sampled segments of the Korean statute corpus.

The overall average scores across the five evaluation dimensions are:  
\textit{Fluency} 4.385, \textit{Practicality} 4.435, \textit{Relevance} 4.540, \textit{Legal Accuracy} 4.515, and \textit{Complexity} 0.915. The agreement between the evaluators also reaches 96\%, with at least two of the three experts assigning consistent labels. Details about inter-annotator agreement are in Appendix~\ref{sec:inter-annotator agreement}. 

While fluency and practicality may involve some degree of subjectivity and are less directly tied to legal reasoning, the consistently high scores in relevance, legal accuracy, and complexity suggest that the dataset effectively captures the demands of factually grounded, multi-hop legal reasoning.
Appendix~\ref{sec:validation_statistics} summarizes the data filtered at each LLM and human validation stage. 

\begin{figure*}[ht]
\begin{center}
   \includegraphics[width=\linewidth]{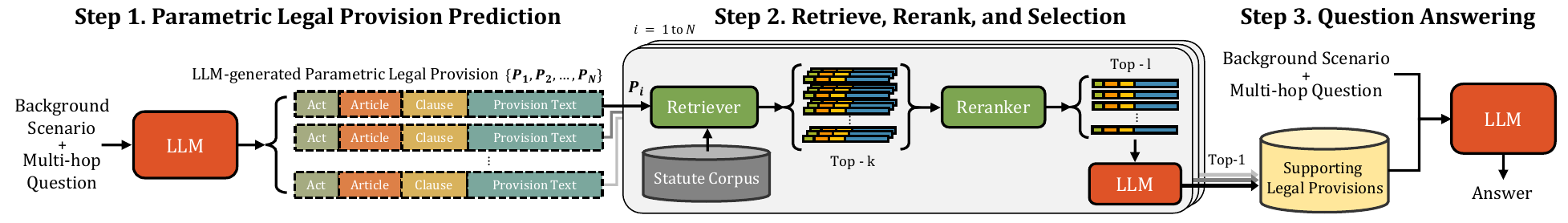}
\end{center}
\vspace{-7pt}
   \caption{\textbf{Illustration of \methodName.} (Step 1) The LLM initially generates parametric provisions. (Step 2) Parametric provisions are then used as queries for \textit{Retrieve, Rerank, and Selection} retrieval. (Step 3) Finally, collected supporting legal provisions are used to multi-hop reasoning for generating responses.}
\label{fig:qa_pipeline}
\end{figure*}

\subsection{English Version of \datasetName}
Unlike previous Korean legal benchmarks \citep{hwang2022multi,kimyeeun-etal-2024-developing}, which only provide Koean data, we release an English-translated version of \datasetName{} to promote broader accessibility and facilitate multilingual research on legal question answering.  
Specifically, $B$, $Q$, and $A$ are translated using GPT-4o, while the statutory provisions $C$ are primarily based on high-quality official translations provided by the Korea Legislation Research Institute. \footnote{\url{https://elaw.klri.re.kr/eng_service/main.do}} 
A small subset of provisions in $C$ that are not covered by the official source are translated via GPT-4o and marked with the tag \textit{\%MACHINE\_TRANSLATED\%}.
\section{\methodName}
We introduce Parametric provision guided Selection Retrieval (\methodName), a method designed to retrieve supporting legal provisions for complex legal questions effectively.

\subsection{Parametric Provision Generation}
Complex legal questions ($Q$) often require reasoning over multiple statutory provisions. 
However, relying solely on the parametric knowledge of LLMs can be unreliable, as they are prone to hallucinations and may fail to recall the exact legal texts \cite{dahl2024large}. 
To address this, we incorporate retrieval over an actual statute corpus to ground the model's responses in authoritative legal text.
As illustrated in Step 1 of Figure \ref{fig:qa_pipeline},
we instruct the LLM to generate a set of potentially relevant provisions  ($\{{p_n}_{n=1}^N\}$) from multi-hop legal question ($Q$), where $N$ is the number of generated parametric provisions.
Each $p_n$ reflects a distinct statutory component that may support reasoning over $Q$.
Since $\{{p_n}_{n=1}^N\}$ are generated solely based on the LLM's parametric knowledge, we refer to them as parametric provisions and use them only as intermediate queries. 
This allows us to leverage retrieval based on queries that resemble the target question.

\subsection{Retrieve, Rerank and Selection}
To improve retrieval accuracy, we propose a novel three-stage retrieval approach incorporating Bi-encoder~\cite{huang2013learning} retrieval, Cross-encoder~\cite{reimers-gurevych-2019-sentence} reranking, and selection via an LLM.
As illustrated in Step 2 of Figure~\ref{fig:qa_pipeline}, each parametric provision $p_n$ is used to retrieve the top $k$ most relevant statutory provisions ($Top-k$) from the corpus via a Bi-encoder retriever based on cosine similarity.
The $Top-k$ provisions retrieved for each $p_n$ are reranked using a Cross-encoder reranker to better capture fine-grained relevance.
After reranking, we select $Top-l, ~(l<k)$ provisions from $Top-k$ provisions.
Then, we instruct an LLM to select the most relevant one among the $Top-l$ for each $p_n$. This process enables the collection of reliable supporting legal provisions by leveraging generated parametric provisions.
Finally, the collected supporting legal provisions are fed into the LLM to solve the complex multi-hop legal question through provision-grounded reasoning.
The parametric provisions generated in the initial stage facilitate multi-hop reasoning by enabling \methodName{} to search for each piece of supporting provision.

\section{\evalName}
\begin{figure}[t]
\small
\resizebox{\columnwidth}{!}{%
\begin{tabular}{p{\columnwidth}}  
\toprule
\textbf{Legal Fidelity Evaluation Prompt} \\ \midrule
\textbf{<Task Description>} \\ 
You will be given a complex legal question along with the relevant legal provisions that can be used to resolve it. \\ \vspace{0.05em}
\textbf{<Evaluation Criteria>} \\
Evaluate the legal accuracy of the response on a scale 1 $\sim$ 10. \\ \vspace{0.05em}
\textbf{<Evaluation Steps>} \\
1. Check whether the prediction properly answers the question. \\
2. Check whether the prediction contradicts or omits any legal provisions in the context. \\
3. Heavily penalize when the legal conclusion differs in detail from the expected output. \\
4. Heavily penalize if the prediction contradicts or omits any specific elements from the context. \\
5. Heavily penalize responses that include statements like 'The given context does not include the answer, but generally'. \\ \vspace{0.05em} \\
\textbf{<Query>} \\
Question: \colorbox{gray!30}{\{question\}} ;
Context: \colorbox{gray!30}{\{context\}} \\
Expected output: \colorbox{gray!30}{\{answer\}} ;
Prediction: \colorbox{gray!30}{\{prediction\}} \\
\bottomrule
\end{tabular}
}
\caption{\textbf{Prompt of Legal Fidelity Evaluation (\evalName{}).} \colorbox{gray!30}{\{placeholder\}} indicates a slot to be filled with the corresponding value for evaluation.}
\label{fig:eval}
\vspace{-7pt}
\end{figure}

\datasetName\ is a benchmark designed to evaluate the legal fidelity of responses generated by LLMs. 
However, existing evaluation metrics often fail to assess the legal correctness of these responses \cite{trautmann-etal-2024-measuring}. 
To address this, we introduce Legal Fidelity Evaluation (\evalName), an evaluation framework for measuring legal fidelity.

\begin{table*}[ht]
\centering

\resizebox{\linewidth}{!}{%
\begin{tabular}{@{}ccccccccccccc@{}}
\toprule
                  & \multicolumn{3}{c}{\cellcolor[HTML]{ECF4FF}F-1} & \multicolumn{3}{c}{\cellcolor[HTML]{ECF4FF}EM} & \multicolumn{3}{c}{\cellcolor[HTML]{FFFC9E}Token F-1} & \multicolumn{3}{c}{\cellcolor[HTML]{FFFC9E}\evalName} \\ \midrule
                  & Qwen         & EXAONE    & GPT-4o   & Qwen         & EXAONE   &  GPT-4o   & Qwen           & EXAONE      &  GPT-4o     & Qwen         & EXAONE    &  GPT-4o   \\ \midrule
\small SP$^{\diamondsuit}$ \cite{NEURIPS2020_1457c0d6}                & -                 & -                & -        & -                 & -               & -        & 30.52               & 23.04              & 36.20      & 40.74             & 49.46            & 55.00    \\
\small CoT$^{\diamondsuit}$ \cite{NEURIPS2022_9d560961}               & -                 & -                & -        & -                 & -               & -        & 26.37               & 23.71              & 32.34      & 39.41             & 37.42            & 52.75    \\
\small SP \cite{NEURIPS2020_1457c0d6}  + OR$^{\heartsuit}$             & 21.50             & 21.50            & 21.50    & 7.08              & 7.08            & 7.08     & 32.18               & 20.58              & 26.75      & 45.36             & 43.98            & 36.45    \\
\small CoT \cite{NEURIPS2022_9d560961} + OR$^{\heartsuit}$            & 21.50             & 21.50            & 21.50    & 7.08              & 7.08            & 7.08     & 28.39               & 28.82              & 30.68      & 42.10             & 47.26            & 46.42    \\
\small Self-Ask$^{\spadesuit}$ \cite{press-etal-2023-measuring}          & 9.29              & 8.55             & 8.55     & 2.65              & 1.33            & 1.77     & 16.59               & 14.72              & 7.82       & 34.14             & 37.77            & 22.44    \\
\small IRCoT$^{\spadesuit}$ \cite{trivedi-etal-2023-interleaving}             & 20.42             & 15.89            & 23.91    & 4.42              & 1.77            & 4.42     & 31.62               & 26.31              & 31.39      & 46.78             & 48.20            & 46.68    \\
\small FLARE$^{\spadesuit}$ \cite{jiang-etal-2023-active}             & 40.64                 & 25.23                & 31.75    & 3.98                 & 14.16               & 4.42     & 29.54                   & 21.31                  & 34.37      & 53.76                 & 34.66                & 50.55    \\
\small ProbTree$^{\clubsuit}$ \cite{cao-etal-2023-probabilistic}          & 15.84             & 11.61            & 17.32        & 2.65              & 2.21            & 3.98        & 28.38               & 24.67              & 33.91          & 43.77             & 46.74            & 52.62        \\ 
\small BeamAggr$^{\clubsuit}$ \cite{chu-etal-2024-beamaggr}          & 14.05             & 10.83            & 16.89        & 2.65              & 0.44            & 3.54        & 16.02               & 10.83              & 22.83          & 32.31             &  31.46           & 41.59        \\ 
\midrule
\small \methodName$^{\clubsuit}$ (Ours) & \textbf{46.24}    & \textbf{48.73}            & \textbf{59.41}    & \textbf{17.70}    & \textbf{17.70}           & \textbf{26.99}    & \textbf{40.65}      & \textbf{31.09}              & \textbf{46.14}      & \textbf{56.00}    & \textbf{57.58}            & \textbf{67.26}    \\ \bottomrule
\end{tabular}%
}
\caption{\textbf{Experimental results of various baseline methods on ~\datasetName.} Columns shaded in \colorbox[HTML]{ECF4FF}{blue} measure retrieval accuracy, and columns shaded in \colorbox[HTML]{FFFC9E}{yellow} measure generation accuracy. The best results are highlighted in bold. We utilize Qwen3-32B~\cite{qwen3}, EXAONE-3.5-32B~\cite{exaone-3.5}, and GPT-4o~\cite{openai2024gpt4technicalreport}.
($\diamondsuit$: No-retrieval, $\heartsuit$: One-time retrieval, $\spadesuit$: Iterative retrieval, $\clubsuit$: Sub-query retrieval).}
\label{tab:main1}
\vspace{-7pt}
\end{table*}

\evalName{} builds upon G-Eval \cite{liu-etal-2023-g}, a representative LLM-as-a-Judge evaluation approach. 
It evaluates the legal fidelity of a response by comparing it against the reference legal provisions and the expected answer. 
Figure~\ref{fig:eval} illustrates the prompt used in \evalName{}.
Under the task description of answering legal questions, \evalName{} employs a robust LLM judge to assign a score on a 1–10 scale. The evaluation follows five clearly defined steps, each corresponding to specific criteria: \textit{Answer Relevance}, \textit{Legal Consistency}, \textit{Conclusion Accuracy}, \textit{Context Fidelity}, and \textit{Avoid Generic Responses}. Finally, \evalName{} produces both a scalar score and detailed justifications aligned with each of the five evaluation steps. These suggest that \evalName{} not only provides a reliable score, but also offers interpretable explanations.

To assess \evalName{}'s reliability, we conduct a human evaluation study on generated responses from \datasetName{}, using two independent annotator groups.
The results show that \evalName{} achieves a Pearson correlation of 84.90 with human judgments, outperforming existing evaluation metrics.
Examples and details of the human evaluation study are provided in the Appendix \ref{sec:appendix_LFEval_correlation}.

\section{Experiments} \vspace{-3pt}
We describe our experimental setup, including models, evaluation metrics, retrieval setting, and baselines.
Additional details are in Appendix~\ref{sec:appendix_experiment}.

\subsection{Models}  \vspace{-3pt}
We employ three different LLMs: Qwen3 \cite{qwen3}, EXAONE-3.5 \cite{exaone-3.5}, and GPT-4o \cite{openai2024gpt4technicalreport}. 
These models are selected for their strong Korean language understanding capabilities.
We use BM-25~\cite{bm25} as the retriever and BGE~\cite{chen2024bge}, finetuned on the Korean dataset, as the reranker.

\subsection{Metrics} \vspace{-3pt}
Since \datasetName{} includes gold-supporting provisions and answers, we employ retrieval and generation metrics for comprehensive evaluation.
For retrieval performance, we use Exact Match (EM) and F-1, computed by comparing the retrieved provisions against the gold-supporting provisions. 
For generation performance, we report Token F-1, a standard metric for evaluating free-form QA, and \evalName{} for assessing the legal fidelity.
\vspace{-3pt}
\subsection{Statute Corpus} \vspace{-3pt}
\label{sec:statute_corpus}
We construct a paragraph-level statute corpus to provide fine-grained legal information. 
We include every active statute that has been cited in Korean court decisions between 1998 and 2024\footnote{We obtain documents via the official API (\url{https://open.law.go.kr/}) on March 17, 2025.}.
The final corpus comprises 608 unique statutes, totaling about 233,544 paragraph-level provisions.
We use this statute corpus as a retrieval pool to obtain actual provisions for our experiments.

\begin{figure*}[ht]
    \centering
    \includegraphics[width=\linewidth,]{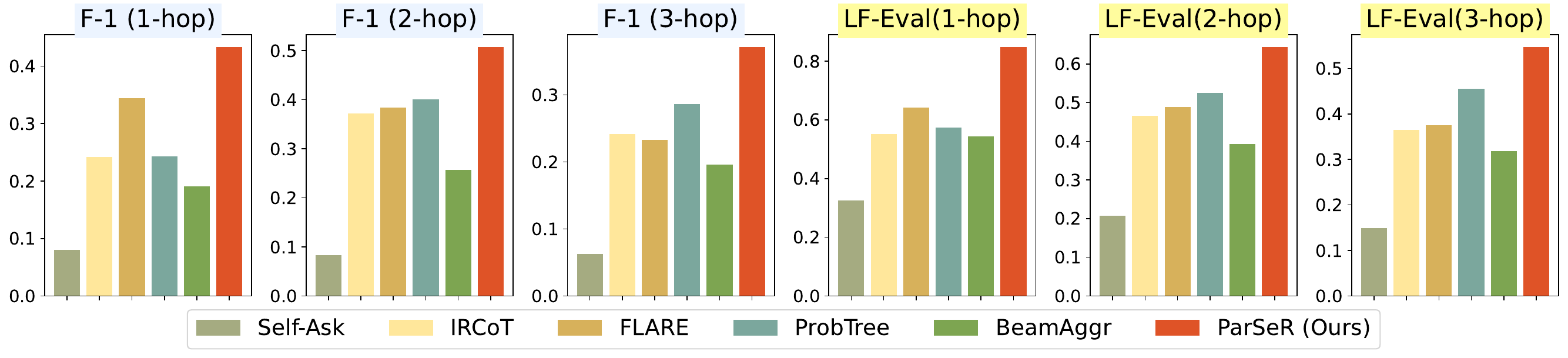}
    \caption{Retrieval performance (F-1) and generation performance (\evalName) of each method by reasoning depth.\\}
    \label{fig:hoplevel}
    \vspace{-7pt}
\end{figure*}
\vspace{-3pt}
\subsection{Baselines} \vspace{-3pt}
To evaluate the effectiveness of \methodName, we compare it against a diverse set of multi-hop reasoning baselines.
Since many baselines target general-domain tasks, official implementations are incompatible for \datasetName{}.
Therefore, we manually re-implemented them for fair comparison. 
We perform retrieval over our statute corpus to ensure consistency for methods requiring retrieval.

We include the following methods as baselines: Standard Prompting (SP) \cite{NEURIPS2020_1457c0d6}, Chain of Thought (CoT) \cite{NEURIPS2022_9d560961}, Self-Ask~\cite{press-etal-2023-measuring}, IRCoT~\cite{trivedi-etal-2023-interleaving}, FLARE~\cite{jiang-etal-2023-active}, ProbTree~\cite{cao-etal-2023-probabilistic}, and BeamAggr~\cite{chu-etal-2024-beamaggr}.
We also include a simple One-Time Retrieval baseline, which gives oracle access to the same number of gold reference provisions as the required number of reasoning hops.
Implementation details about baseline methods are provided in Appendix \ref{appendix:baseline}

\section{ Results} \vspace{-5pt}
Table~\ref{tab:main1} shows experimental results on \datasetName{}, comparing various baseline methods.
We observe that baselines with low F-1 and EM scores tend to show lower Token F-1 and \evalName{} scores than the No-retrieval baseline.\footnote{While most baselines generally improve with iterative reasoning, BeamAggr underperforms because its multi-source retrieval is incompatible with our setup, limiting its advantage.}
This suggests that retrieving irrelevant provisions negatively impacts multi-hop legal reasoning.

On the other hand, \methodName{} consistently outperforms all baseline methods with all LLMs across all retrieval and generation metrics.
Notably, on GPT-4o, \methodName{} significantly surpasses the strongest baseline, ProbTree, with a +12.23 improvement in Token F-1 and +14.64 in \evalName.
This indicates that \methodName{} can effectively perform reasoning over complex multi-hop legal questions.
Additional results on smaller LLMs are in Appendix \ref{smallresults}.

\vspace{-4pt}
\begin{table}[t]
\centering
\resizebox{\columnwidth}{!}{%
\begin{tabular}{@{}ccccc@{}}
\toprule
  & \cellcolor[HTML]{ECF4FF}F-1 & \cellcolor[HTML]{ECF4FF}EM & \cellcolor[HTML]{FFFC9E}Token F-1 & \cellcolor[HTML]{FFFC9E}\evalName \\ \midrule
\methodName\   & \textbf{48.74} & \textbf{17.70}   & \textbf{31.09} & \textbf{57.58} \\ \midrule
\small w/o Selection & 40.61 & 13.72   & 27.11   &  50.18 \\
\small w/o Reranking & 40.64 & 14.16  & 29.54   & 54.02 \\
\small w/o Reranking, Selection & 27.56 & 6.64   &  25.90  & 45.97  
\\
\small w/o Reranking, Selection, Provision & 21.41 & 3.98   &  21.20  & 45.52  
\\\bottomrule
\end{tabular}%
}
\caption{\textbf{Ablation results on \methodName\ on EXAONE.}
"w/o Provision" replaces parametric provision generation with top-$k$ retrieval based on the original question, where $k$ matches the number of generated provisions.
}
\label{tab:ablation}
\vspace{-7pt}
\end{table}
\section{Analyses}

\paragraph{Reasoning Depth.}Given that \datasetName{} encompasses scenario-based QA tasks requiring varying levels of reasoning depth, we conduct a detailed analysis by evaluating performance across different hop levels.
Figure \ref{fig:hoplevel} shows the performances with respect to different reasoning depths.
\methodName{} consistently achieves the best performance across all hop levels.
Excluding \methodName{}, confidence-based methods such as FLARE and ProbTree outperform other baselines.
Specifically, FLARE ranks second at the 1-hop level, while ProbTree excels in deeper reasoning at the 2-hop and 3-hop levels.
These results suggest that the performance of different existing methods can vary depending on the required reasoning depth.
In contrast, \methodName{} demonstrates robust performance across questions with all levels of reasoning depth.

\begin{figure}
    \centering
    \includegraphics[width=\linewidth,height=4.5cm]{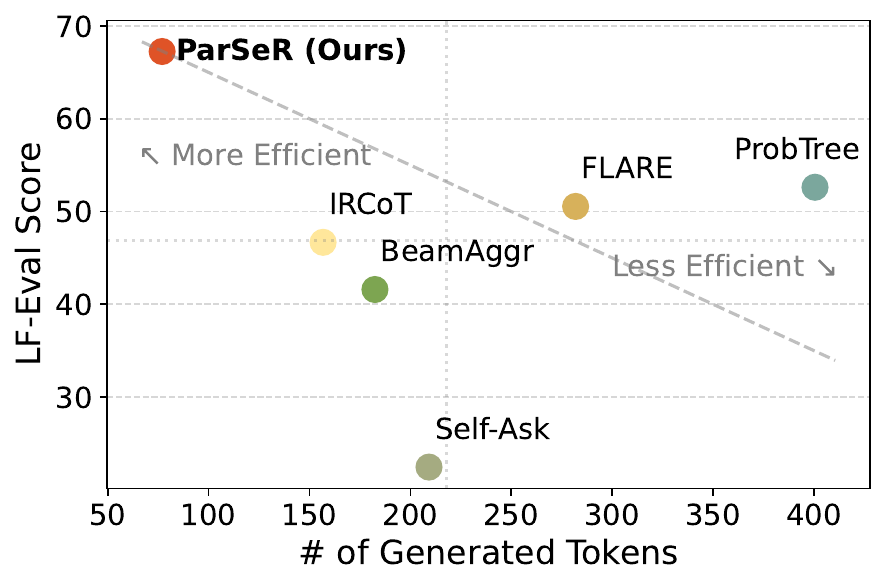}
    \caption{Average \methodName{} scores with respect to average number of generated tokens by GPT-4o.} 
    \label{fig:efficiency}
    \vspace{-7pt}
\end{figure}

\paragraph{Ablation Study.}
To analyze the role of each module in driving the performance improvements of \methodName{}, we conduct an ablation study by isolating the impact of each component.
Table~\ref{tab:ablation} shows the ablation study of our method on EXAONE 3.5-32B \cite{exaone-3.5}.
Removing \textit{Selection} causes a greater performance drop than removing \textit{Reranking}, suggesting that leveraging the LLM’s ability to select relevant provisions is more effective than cross-encoder.
We observe a significant performance drop when both \textit{Reranking} and \textit{Selection} are removed.
Furthermore, replacing \textit{parametric provision} generation with simple top-$k$ retrieval based on the original question, where $k$ equals the number of generated provisions, leads to even worse performance.
This indicates that parametric provision generation better supports multi-hop retrieval-augmented reasoning than simple retrieval.

\paragraph{Efficiency.}
While retrieval-augmented reasoning methods significantly improve performance on complex questions, they often require a large number of tokens for final answer prediction.
To assess both effectiveness and efficiency, we analyze the performance of each method relative to the number of tokens consumed during answer prediction.
Figure~\ref{fig:efficiency} presents \evalName{} score against the average number of generated tokens.
While methods with more generated tokens tend to show increased performance, the improvements are generally marginal relative to the computational overhead, suggesting potential inefficiency.
However, \methodName{} demonstrates the highest performance while generating the fewest tokens, making it the most efficient approach.
Unlike other baselines, \methodName{}  leverages parametric provisions generated in the initial stage to facilitate targeted multi-hop evidence retrieval while focusing on improving retrieval accuracy through a 3-stage retrieval pipeline.
This feature leads that \methodName{} is not only effective but also cost-efficient for multi-hop legal reasoning.
\vspace{-4pt}
\begin{table}[t]
\centering
\resizebox{\columnwidth}{!}{%
\begin{tabular}{@{}ccccc@{}}
\toprule
  & \cellcolor[HTML]{ECF4FF}F-1 & \cellcolor[HTML]{ECF4FF}EM & \cellcolor[HTML]{FFFC9E}Token F-1 & \cellcolor[HTML]{FFFC9E}\evalName \\ \midrule
\methodName\ (Sparse)   & 48.74 & \textbf{17.70}   & \textbf{31.09} & \textbf{57.58} \\ 
\methodName\ (Dense)   & \textbf{50.43} & 16.81   & 29.56 & \textbf{57.58} \\ 
\methodName\ (Hybrid)   & 48.16 & 16.81   & 28.76 & 56.61 \\
\bottomrule
\end{tabular}%
}
\caption{\textbf{Performance of \methodName{} across different retriever types.} We compare sparse (BM25~\cite{bm25}), dense (BGE-M3~\cite{chen2024bge}), and hybrid (BGE-M3 hybrid) retrievers using EXAONE 3.5-32B~\cite{exaone-3.5}.
}
\label{tab:ret_type}
\vspace{-7pt}
\end{table}

\paragraph{Effect of Retreiver Type.} Since various types of retrieval tools are available, we investigate the impact of retriever type on the performance of \methodName{}.
For the legal provision retrieval task, we consider BM25~\cite{bm25} as a sparse retriever, BGE-M3~\cite{chen2024bge}, known for its strong Korean embedding capabilities, as a dense retriever, and their combination as a hybrid retriever.
Table~\ref{tab:ret_type} presents the experimental results of using different retriever types with EXAONE-3.5-32B.
Overall, there is no significant performance difference across retriever types, although the sparse retriever (BM25) achieves the highest EM, token-level F1, and \evalName{} score.
Therefore, we adopt the sparse retriever (BM25) as the retrieval tool, considering its accessibility for reproducibility, efficient retrieval speed, and fairness in comparison. See Appendix~\ref{appendix:baseline} for a detailed description of the retrieval configuration.

\vspace{-4pt}
\begin{table}[t]
\small
\centering
\resizebox{\columnwidth}{!}{%
\begin{tabular}{@{}ccccc@{}}
\toprule
 $Top-k$ & \cellcolor[HTML]{ECF4FF}F-1 & \cellcolor[HTML]{ECF4FF}EM & \cellcolor[HTML]{FFFC9E}Token F-1 & \cellcolor[HTML]{FFFC9E}\evalName \\ \midrule
\textit{Top-50}   & 47.98 & 16.81   & 30.24 & 58.93 \\ \midrule
\textit{Top-100}   & 48.74 & 17.70   & 31.09 &57.58 \\ \midrule
\textit{Top-200}   & 49.29 & 17.70   & 31.20 & \textbf{60.59} \\
\textit{Top-300}   & \textbf{50.26} & \textbf{18.58}   & \textbf{31.97} & 60.58 \\
\bottomrule
\end{tabular}%
}
\caption{\textbf{Analysis on $k$ value on retrieval.} We vary the $k$ value (50, 100, 200, 300) while keeping other modules of \methodName{} fixed, using EXAONE 3.5-32B.
}
\label{tab:topk}
\vspace{-7pt}
\end{table}
\vspace{-4pt}
\begin{table}[t]
\small
\centering
\resizebox{\columnwidth}{!}{%
\begin{tabular}{@{}ccccc@{}}
\toprule
 $Top-l$ & \cellcolor[HTML]{ECF4FF}F-1 & \cellcolor[HTML]{ECF4FF}EM & \cellcolor[HTML]{FFFC9E}Token F-1 & \cellcolor[HTML]{FFFC9E}\evalName \\ \midrule
\textit{Top-5}  & 47.56 & 17.26   & 30.63 & \textbf{59.21} \\ \midrule
\textit{Top-10}  & 48.74 & 17.70   & \textbf{31.09} &57.58 \\ \midrule
\textit{Top-20}  & 50.03 & 18.14   & 30.67 & 59.14 \\
\textit{Top-30}   & \textbf{50.34} & \textbf{19.03}   & 30.43 & 58.49 \\
\bottomrule
\end{tabular}%
}
\caption{\textbf{Analysis on $l$ value on reranking.} We vary the $l$ value (5, 10, 20, 30) while keeping other modules of \methodName{} fixed, using EXAONE 3.5-32B.
}
\label{tab:topl}
\vspace{-7pt}
\end{table}
\vspace{-7pt}

\paragraph{Effect of $k$ and $l$ in \methodName{}.} We investigate the impact of two key hyperparameters in the 3-stage retrieval pipeline: $k$ (retrieval scope) and $l$ (reranking scope).
Table~\ref{tab:topk} presents the results of varying $k$ while keeping other components of \methodName{} fixed.
We observe a slight improvement in retrieval metrics as $k$ increases, while the effect on generation metrics remains marginal.
Table~\ref{tab:topl} shows the results of varying $l$ under the same conditions. Unlike $k$, increasing $l$ does not lead to consistent performance gains.
While larger $k$ and $l$ values offer broader context and better recall, they also increase computational cost in reranking and risk exceeding context limits during selection.
Therefore, we set $k=100$ and $l=10$ for \methodName{} in this paper.

\vspace{-3pt}
\section{Conclusion} \vspace{-7pt}
In this work, we introduce \textbf{\datasetName}, a benchmark designed to evaluate LLMs on provision-grounded, open-ended legal QA in Korean law. 
Furthermore, we present \textbf{\methodName}, which significantly outperforms existing retrieval-augmented reasoning methods in both retrieval accuracy and answer quality.
To enable reliable assessment of legal fidelity in free-form responses, we propose \textbf{\evalName}, an automatic evaluation aligned with human judgment. 

Our experiments demonstrate that \methodName{} consistently outperforms all baselines across all metrics in a cost-efficient and reasoning-depth-agnostic manner.
We hope \datasetName, \evalName, and \methodName{} serve as valuable resources for advancing research in legal NLP.

\section*{Limitations}
\paragraph{Limited Scale and Expert Dependency.}
The current version of \datasetName\ is relatively limited in size. While our automated pipeline enables the generation of initial QA drafts at scale, each instance still requires careful review and revision by legal experts to ensure legal correctness and contextual coherence. This expert validation step remains essential given the current limitations of LLMs in reliably handling nuanced legal interpretation without human oversight.

\paragraph{Civil Law Focus.}
\datasetName, \methodName, and \evalName{} are developed based on Korean statutory law, reflecting the characteristics of a civil law system where codified statutes serve as the primary source of legal authority. While this design enables rigorous evaluation of statute-grounded legal reasoning, it may limit the applicability to common law jurisdictions, such as those in the United States or the United Kingdom, where legal interpretation heavily relies on case law and judicial precedents.

\section*{Ethical Considerations}
To ensure ethical data construction and usage, several precautions are taken during the development of \datasetName.

\paragraph{Dataset Construction.}
First, all question, answer instances in the dataset are derived from fictional legal scenarios and do not contain any personally identifiable information. Every background scenario is composed using anonymized character names (e.g., Person A, Person B), and no real individuals, cases, or sensitive details are included. 

Second, all legal experts who participated in the revision and evaluation process were informed in advance that the purpose of the task is to construct a dataset for evaluating LLM performance on legal reasoning (see Appendix~\ref{sec:appendix_annotator_guidelines}).

Finally, all legal documents used in constructing \datasetName—both statutes and precedents—were collected from official government APIs and fall under Korea's public data policy, which permits their redistribution for research purposes. For the English versions of the statutes, we used publicly released translations by the Korea Legislation Research Institute (KLRI), which confirms redistribution is allowed for non-commercial research use. Any machine-translated content is labeled within the dataset to maintain transparency (Appendix~\ref{sec:appendix_license}).

\paragraph{Intended Use.} This work presents a benchmark and methodology for evaluating open-ended, provision-grounded legal question answering in Korean. It is designed solely for research purposes and is not intended for direct use in real-world legal decision-making or as a substitute for professional legal advice. The benchmark aims to support the development and evaluation of legal NLP systems.

\section*{Acknowledgments}
This research was supported by the MSIT(Ministry of Science and ICT), Korea, under the ITRC(Information Technology Research Center) support program(IITP-2025-RS-2020-II201789) supervised by the IITP(Institute for Information \& Communications Technology Planning \& Evaluation, Contribution Rate: 47.5\%). This research was supported by Culture, Sports and Tourism R\&D Program through the Korea Creative Content Agency grant funded by the Ministry of Culture, Sports and Tourism in 2025 (Project Name: Development of an AI-Based Korean Diagnostic System for Efficient Korean Speaking Learning by Foreigners, Project Number: RS-2025-02413038, Contribution Rate: 47.5\%). This work was also supported by the Institute of Information \& Communications Technology Planning \& Evaluation (IITP) grant funded by the Korea government (MSIT) (No. RS-2019-II191906, Artificial Intelligence Graduate School Program (POSTECH), Contribution Rate: 5\%).

\bibliography{custom}

\clearpage
\appendix
\onecolumn
\section{Data Filtering and Curation}
\subsection{Final Inclusion Criteria}
\label{sec:dataset_filtering_criteria}
During the human expert revision phase, each QA instance is categorized as either \textit{Pass}, \textit{Revise}, or \textit{Hold}.  
Following this revision phase, all instances are evaluated by three legal experts according to the five evaluation criteria described in Appendix~\ref{sec:appendix_annotator_guidelines}.
Instances labeled as \textit{Pass} or \textit{Revise} are included in the final dataset if they satisfy both of the following scoring conditions during the evaluation phase:
\begin{itemize}
    \item \textbf{Average score threshold:}  
    The instance has to receive an average score of at least 3.0 from the three annotators for each of the following criteria: \textbf{Fluency}, \textbf{Practicality}, \textbf{Relevance}, and \textbf{Legal Accuracy}.

    \item \textbf{Complexity threshold:} At least two of the three annotators have to assign a complexity as True, indicating that all provided legal texts are required to solve the question.
\end{itemize}

A subset of samples initially marked as \textit{Hold}—specifically, those for which the evaluators indicated that the issue could be resolved by adding additional statutory provisions—is also sent to the evaluation stage with the necessary legal context attached.  
As with the \textit{Pass}/\textit{Revise} group, only samples that meet the above evaluation criteria are retained.

All other \textit{Hold} cases are excluded from the final benchmark.  
This included instances where interpretation requires not only statutory provisions but also precedent-based reasoning to reach a sound legal conclusion or scenarios that are unrelated to the provided legal provisions.

Instances that fail to meet the evaluation criteria specified above are likewise excluded from the final release regardless of their initial label.

\subsection{Validation Filtering Statistics}
\label{sec:validation_statistics}
\begin{table}[ht]
\centering
\resizebox{\columnwidth}{!}{%
\begin{tabular}{lrrrrr}
 \hline
 & \textbf{1-hop} & \textbf{2-hop} & \textbf{3-hop} & \textbf{4-hop} & \textbf{Total} \\ \hline
\textbf{Step (2) After Reformulation} & 648 & 677 & 1353 & 357 & 3035 \\ \hline
- A. Full Check & 636 (98\%) & 647 (96\%) & 1273 (94\%) & 337 (94\%) & 2893 (95\%) \\
- B. Partial Check & 614 (95\%) & 164 (24\%) & 58 (4\%) & 4 (1\%) & 840 (28\%) \\
\textbf{Step (3) After LLM Val. (A $\cap$ B)} & 601 (93\%) & 160 (24\%) & 51 (4\%) & 4 (1\%) & 816 (27\%) \\ \hline
\textbf{* Before Human Val.} & 67 (10\%) & 158 (23\%) & 51 (4\%) & 4 (1\%) & 280 (9\%) \\ \hline
\textbf{Step (4\&5) After Human Val.} & 55 (8\%) & 125 (18\%) & 46 (4\%) & 0 (0\%) & 226 (7\%) \\ \hline
\end{tabular}
}
\caption{
Number of QA instances before and after each validation step.  
\textbf{Step (2)} refers to the number of QA instances after scenario-based reformulation (Section~\ref{sec:scenario_reformulation}).  
\textbf{Step (3)} shows the number of instances that passed both the Partial Check and Full Check during the LLM validation stage (Section~\ref{sec:llm_validation}).  
To maintain a balanced distribution across reasoning levels, a subset of 1-hop and 2-hop instances is selectively excluded before the human validation phase (\textbf{Before Human Validation}).  
\textbf{Step (4\&5)} presents the final number of instances that passed both human revision and evaluation (Section~\ref{sec:human_eval}).
Percentages represent the pass rate relative to the number of instances after scenario-based reformulation.
}
\label{tab:filtering_statistics_total}
\end{table}
Table~\ref{tab:filtering_statistics_total} presents the number of QA instances retained and filtered across each validation stage of the benchmark construction pipeline.

After scenario-based reformulation (Step 2), a total of 3,035 QA drafts are produced.
During the LLM validation stage (Step 3), only 816 instances (27\%) pass both the Partial and Full Checks. 

The pass rate for the Partial Check drops sharply as the reasoning level increases—from 95\% for 1-hop questions to just 1\% for 4-hop—indicating that constructing valid multi-hop questions becomes substantially more difficult as reasoning depth increases.
To compensate for this difficulty and ensure sufficient coverage at higher reasoning levels, we generate and validate additional 3-hop samples.

To maintain a balanced distribution across reasoning levels, we selectively exclude a portion of 1-hop and 2-hop instances prior to human validation (marked as * in the table).

Following human expert revision and evaluation (Step 4\&5), 226 QA instances remain in the final dataset.
These include 55 single-hop, 125 two-hop, and 46 three-hop questions, with all 4-hop questions removed during the final filtering process described in Appendix \ref{sec:dataset_filtering_criteria}.

\begin{figure}
\begin{center}
   \includegraphics[width=1\linewidth]{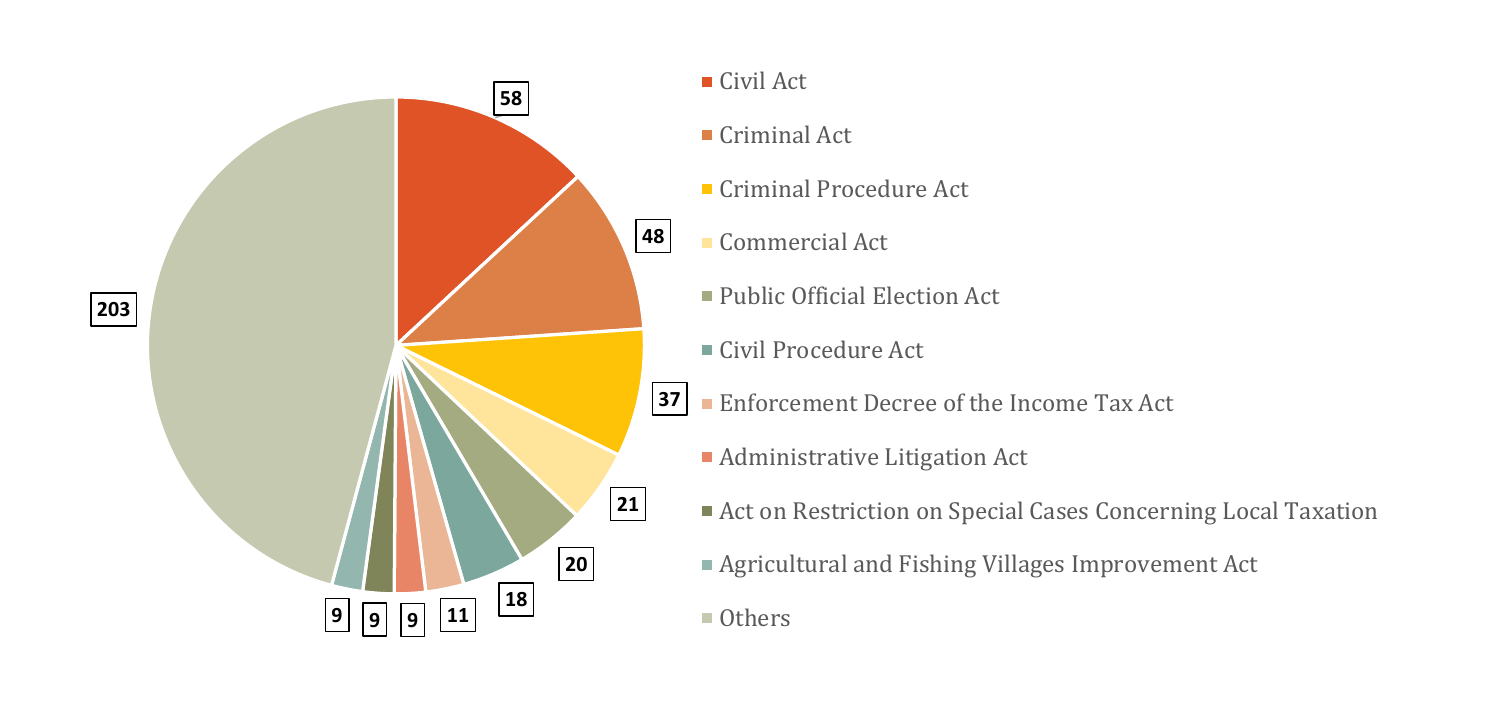}
\end{center}
   \caption{\textbf{Distribution of Statutes in \datasetName.}  
shows the most frequently appearing statutes in the benchmark—specifically, those cited in at least nine QA instances.  
Statutes referenced fewer than nine times are aggregated into the \textit{Others} category for clarity.  
The numbers outside each segment indicate the number of QA instances associated with each statute.}
\label{fig:statute_distribution}
\end{figure}

\section{List and Distribution of Statutes}
\label{sec:appendix_distribution_statutes}
Figure~\ref{fig:statute_distribution} visualizes the distribution of statutes cited in \datasetName.  
For readability, only statutes that appear in at least nine QA instances are shown individually.  
Statutes cited fewer than nine times are aggregated into the \textit{Others} category.  
The complete list of all 83 statutes and their corresponding frequencies is included in the released benchmark.

\section{Licensing}
\label{sec:appendix_license}
\subsection{Source Data Licensing and Usage Rights}
We utilized the public API provided by the \textit{Korea Ministry of Government Legislation’s Law Information Sharing Service}.\footnote{\url{https://open.law.go.kr/}}  
This API was used to collect both statutory provisions and precedents for the construction of our benchmark.

According to the \textit{Act on Promotion of the Provision and Use of Public Data}, all legal information provided by the Korean Law Information Center—excluding English translations—is classified as public data.  
This information is openly accessible and may be freely used, including for commercial purposes, without restriction.  
Accordingly, our use of these sources is fully compliant with the relevant licensing and usage policies.

For the English versions of Korean statutes, we used translations provided by the Korea Legislation Research Institute (KLRI).\footnote{\url{https://elaw.klri.re.kr/eng_service/main.do}}  
We confirmed with the KLRI that these translations may be redistributed for research purposes, provided that proper attribution is given and the usage remains non-commercial. 

For statutes that are not covered by KLRI’s official English translations—approximately 22 provisions—we used machine translation to generate their English versions.
These machine-translated segments are explicitly marked in the dataset with the tag \textit{\%MACHINE\_TRANSLATED\%} to ensure transparency.

\subsection{\datasetName\ License}

The \datasetName\ benchmark is released under the \textbf{Creative Commons Attribution-NonCommercial 4.0 International (CC BY-NC 4.0)} license.\footnote{\url{https://creativecommons.org/licenses/by-nc/4.0/}}  
This license permits users to copy, modify, and redistribute the dataset for non-commercial purposes as long as appropriate credit is attributed to the original authors.  
Any commercial use of the benchmark or its derivatives is strictly prohibited without prior written permission from the authors.

\subsection{Licenses of artifacts} 
The EXAONE is licensed under the EXAONE AI Model License Agreement 1.1 - NC, which permits non-commercial research use only. 
See \href{https://huggingface.co/LGAI-EXAONE/EXAONE-3.5-32B-Instruct}{EXAONE-License} for details.
Other artifacts employed in this research are publicly available.
\section{Inter-Annotator Agreement Analysis}
\label{sec:inter-annotator agreement}

\begin{figure}[ht]
\begin{center}
   \includegraphics[width=0.8\linewidth]{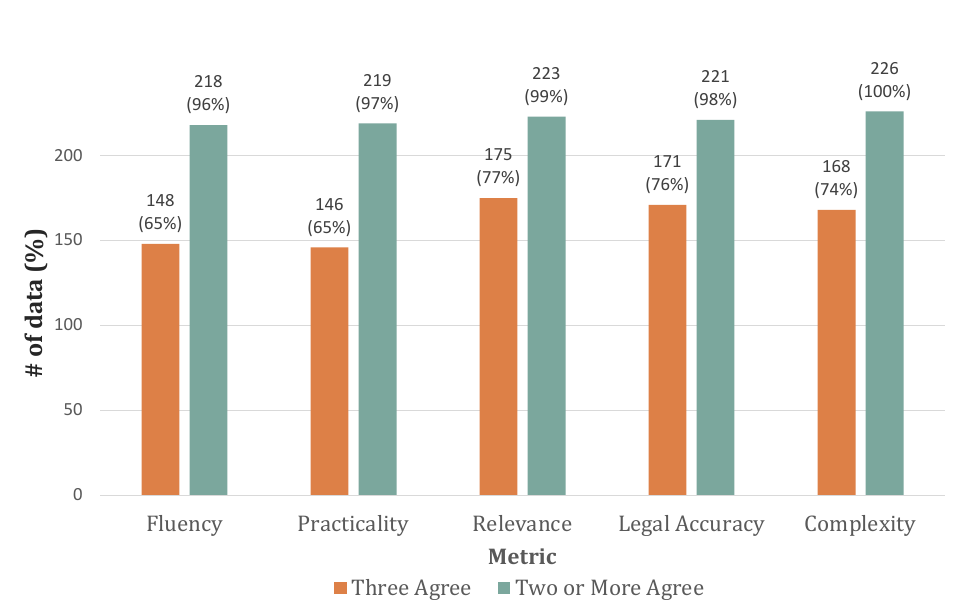}
\end{center}
   \caption{\textbf{Inter-annotator agreement for each evaluation metric.} We report the number and proportion of QA instances where at least two annotators agreed (\textit{Two or More Agree}, blue) and where all three annotators provided identical labels (\textit{Three Agree}, orange). Labels are categorized as positive (Likert scores 4–5), neutral (3), or negative (1–2).}
\label{fig:inter_annotator}
\end{figure}

As described in Section \ref{sec:human_eval}, three legal experts independently evaluate each revised QA instance. Fluency, practicality, relevance, and legal accuracy are rated on a 5-point Likert scale (1 to 5), while complexity is assessed as a binary label (0 or 1).

To enable agreement analysis, we re-map the Likert-scale scores into three ordinal categories, following the methodology of \citet{fan-etal-2019-eli5}:

\begin{itemize}
\item \textbf{Positive}: scores of 4 or 5
\item \textbf{Neutral}: score of 3
\item \textbf{Negative}: scores of 1 or 2
\end{itemize}

Figure \ref{fig:inter_annotator} presents the agreement levels across the five evaluation criteria.  
For all metrics, over 95\% of the samples exhibit agreement between at least two annotators.  
Moreover, more than 65\% of the instances show full agreement across all three annotators, indicating a high level of reliability in the expert judgments.
\section{Guidelines for QA Construction}
\label{sec:appendix_annotator_guidelines}
This section includes detailed guidelines distributed to legal experts for revising and evaluating QA instances, as well as information on the compensation provided to annotators. 

\subsection{Revision Guidelines for QA Drafts}
Each legal QA instance in our dataset consists of the following four components:

\begin{itemize}
    \item \textbf{Background Scenario}: A narrative description of a specific legal situation or dispute.
    \item \textbf{Question}: A legal issue that naturally arises from the given scenario.
    \item \textbf{Answer}: A concise and logically sound response derived strictly from the provided statutes.
    \item \textbf{Reference Provisions}: Legal provisions that serve as the sole basis for answering the question.
\end{itemize}

Legal experts are instructed to revise each instance to ensure legal soundness and linguistic naturalness. 
The resulting dataset is intended to evaluate how well large language models can understand and reason in legal contexts.
The revision is conducted to satisfy the following five criteria:

\begin{itemize}
    \item \textbf{Fluency}: Sentences must be grammatically correct, coherent, and natural.
    \item \textbf{Practicality}: Questions should move beyond simple definitions and be framed to support practical legal reasoning.
    \item \textbf{Relevance}: The background and question must be closely grounded in the provided legal provisions, enabling an answer based solely on them.
    \item \textbf{Legal Accuracy}: Answers must faithfully and correctly interpret the given legal provisions.
    \item \textbf{Complexity}: Resolving the question must require using all provided legal documents without relying on external information.
\end{itemize}

Each instance is labeled with one of the following revision statuses:

\begin{itemize}
    \item \textbf{Pass}: All components are accurate, coherent, and closely tied to the statutes. No revision is necessary.
    \item \textbf{Revise}: Applied when any of the following issues are identified:
    \begin{itemize}
        \item Ungrammatical, awkward, or unnatural phrasing in any component.
        \item Logical inconsistencies, such as mismatched subjects (e.g., confusion between parties) or timeline errors.
        \item Redundant mention of legal article numbers or statutory content in the scenario or question.
        \item Questions that are too general, factual, or fail to articulate a clear legal issue.
        \item Answers that are legally incorrect, overly vague, or lack logical coherence.
        \item Statutes mentioned without clearly naming the corresponding law.
    \end{itemize}
    \item \textbf{Hold}: Used when meaningful revision is not feasible, including:
    \begin{itemize}
        \item Extremely low-quality inputs or incoherent writing.
        \item Questions that require external legal knowledge or unstated assumptions.
        \item Subjective or factual questions that do not require legal interpretation.
        \item Scenarios or questions that are unrelated to the provided legal provisions.
    \end{itemize}
\end{itemize}

\textbf{Important:} All revisions must strictly adhere to the given legal statutes. Even if a different interpretation may apply in actual legal practice, annotators are instructed to provide and revise answers solely based on the scope of the provided provisions.

\subsection{Evaluation Guidelines for QA Instances}
This section describes the evaluation guideline used to assess the quality of the revised legal QA pairs.
Evaluators are informed that the resulting dataset would be used to evaluate the legal reasoning capabilities of large language models.
Each instance, previously corrected by an expert, consists of four components:

\begin{itemize}
    \item \textbf{Background Scenario}: A narrative description of a specific legal situation or dispute.
    \item \textbf{Question}: A legal issue that naturally arises from the given scenario.
    \item \textbf{Answer}: A concise and logically sound response derived strictly from the provided statutes.
    \item \textbf{Reference Provisions}: Legal provisions that serve as the sole basis for answering the question.
\end{itemize}

Evaluators are instructed to assess each QA pair based on the following five criteria. Each criterion is scored independently on a 5-point Likert scale: \textit{Excellent}, \textit{Good}, \textit{Fair}, \textit{Poor}, and \textit{Very Poor}.

The evaluation is conducted according to the following dimensions:

\begin{itemize}
    \item \textbf{Fluency}: Sentences must be grammatically correct, coherent, and natural.
    \item \textbf{Practicality}: Questions should move beyond simple definitions and be framed to support practical legal reasoning.
    \begin{itemize}
        \item E.g., generic questions like ``What is the definition of a car?'' should be rated low in practicality.
    \end{itemize}
    \item \textbf{Relevance}: The background and question must be closely grounded in the provided legal provisions, enabling an answer based solely on them.
    \item \textbf{Legal Accuracy}: Answers must faithfully and correctly interpret the given legal provisions.
    \begin{itemize}
        \item \textbf{Important:} This must be evaluated strictly based on the provided legal texts.
    \end{itemize}
    \item \textbf{Complexity}: Resolving the question must require using all provided legal documents without relying on external information.
    \begin{itemize}
        \item This is a multiple-choice evaluation item. Evaluators must select all legal provisions that are necessary to answer the question accurately. It is important to identify all reference provisions without omission. 
        \item If none of the provided statutes are necessary, or if additional legal provisions are needed to answer the question, evaluators should explicitly indicate this.
    \end{itemize}
    
\end{itemize}
\subsection{Compensation for Legal Expert Annotation}
\label{sec:appendix_cost}
Each legal expert annotator is compensated with a stipend of 100,000 KRW (approximately \$75 USD) for completing 80 QA instances, consisting of 20 items for revision and 60 for evaluation.
Based on our internal task estimation, this amount corresponds to approximately 2.5 hours of expert work, yielding an effective hourly wage of 40,000 KRW. This rate is roughly four times the Korean legal minimum wage (as of 2025) and is deliberately set to ensure fair compensation and to attract high-quality legal annotators.

\section{Dataset Construction Prompt Templates}
\label{sec:appendix_prompt_templates}
Figures \ref{fig:prompt_qg}--\ref{fig:prompt_llm_validation} present the prompt templates used with GPT-4o during the question \& answer generation and two-stage validation process.
Figure~\ref{fig:prompt_qg} shows the prompts used for question \& answer generation (Section~\ref{sec:qa_construction});  
Figure~\ref{fig:prompt_scenario_reformulation} displays the prompt used for scenario-based reformulation (Section~\ref{sec:scenario_reformulation});  
and Figure~\ref{fig:prompt_llm_validation} presents the prompts used for LLM validation (Section~\ref{sec:llm_validation}).
\\

\begin{figure}[ht]
\small
\begin{tabular}{p{\columnwidth}}  
\toprule
\textbf{Part 1. Instruction for Q\&A Generation (1hop):}\\ \midrule
\begin{tabular}[c]{@{}l@{}}You are given a set of legal provisions (in Korean). Your task is to generate a single-hop legal question-answer \\ pair based on the most appropriate provision from the context.\\ \\ Your task is to:\\ 1. Select one provision from the given context that is legally meaningful and suitable for generating a question.\\ 2. Create a clear and legally relevant question that reflects the core content of the selected provision.\\ 3. Provide an accurate legal answer that can be answered solely based on the selected provision.\\ 4. If none of the given provisions are appropriate for generating a question, output only the line: "Not applicable"\\ \\ Output Format:\\ question: {[}Generated legal question{]}  \\ answer: {[}Accurate answer based on the provision{]}  \\ selected\_context: {[}Only include the provision used for the Q\&A{]}\\ \\ Example: \{Example\}\\\\ \textbf{<Query>} \\ Context: \colorbox{gray!30}{\{context\}}\end{tabular} \\ \midrule
\textbf{Part 2. Instruction for Q\&A Generation (mhop):} \\ \midrule
\begin{tabular}[c]{@{}l@{}}You are given:\\ An existing Q\&A pair generated from one or more legal provisions.\\ \\ 1. The current\_context, which contains the provision(s) used for the existing Q\&A.\\ 2. The remain\_context, which contains additional legal provisions not yet used.\\ 3. Your task is to expand the original question into a (k+1)-hop question that logically incorporates one new \\ provision from remain\_context.\\ \\ Your task is to:\\ 1. Select one provision from remain\_context that logically connects to the existing question or answer.\\ 2. Generate an expanded (k+1)-hop legal question that requires both current\_context and the newly selected \\ provision to be answered.\\ 3. Provide a new answer that integrates both contexts.\\ 4. Output the newly selected provision as selected\_context.\\ 5. If none of the remaining provisions are appropriate for expansion, output only the line: "Not applicable"\\ \\ Constraints\\ 1. The new question must logically build upon the existing question.\\ 2. The answer must not be answerable using only one of the contexts \\
(neither current\_context nor the selected\_context alone).\\ 3. The question must be in Korean, concise, and naturally phrased.\\ \\ Output Format:\\ question: {[}Expanded k+1-hop legal question{]}  \\ answer: {[}New answer that depends on both current\_context and selected\_context{]}  \\ selected\_context: {[}One newly selected sentence from remain\_context{]}\\ \\ Example: \{Example\}\\\\\textbf{<Query>} \\ Question: \colorbox{gray!30}{\{question\}}\\ Answer: \colorbox{gray!30}{\{answer\}}\\ Current\_context: \colorbox{gray!30}{\{current\_context\}}\\ Remain\_context: \colorbox{gray!30}{\{remain\_context\}}\end{tabular} \\ \bottomrule
\end{tabular}
\caption{Prompt templates used for generating legal Q\&A pairs from statutory text. Part 1 describes instructions for single-hop question generation based on a single legal provision, while Part 2 outlines multi-hop question generation that requires reasoning over multiple provisions. \colorbox{gray!30}{\{placeholder\}} indicates a slot to be filled with the corresponding value.}
\label{fig:prompt_qg}
\end{figure}
\begin{figure}[ht]
\small
\begin{tabular}{p{\columnwidth}}  
\toprule
\textbf{Part 3. Instruction for Scenario-based reformulation}\\ \midrule
\begin{tabular}[c]{@{}l@{}}You are given a legal question-answer pair based on statutory interpretation. \\ Rewrite this QA pair into a realistic legal scenario involving fictional characters (e.g., Person A, Person B) \\ where the same legal logic would apply.\\ \\ Do the following:\\ 1. Create a short but concrete fact pattern (case scenario) that would require applying the same legal reasoning.\\ 2. Rewrite the original question to match the scenario.\\ 3. Keep the original legal answer, with minor edits if needed to match the scenario.\\ \\ Output Format:\\ background\_scenario: \{BACKGROUND SCENARIO\}\\ question: \{MULTI-HOP LEGAL QUESTION\}\\ answer: \{ANSWER\}\\\\ \textbf{<Query>} \\
Question: \colorbox{gray!30}{\{question\}}\\ Answer: \colorbox{gray!30}{\{answer\}}\\ Context: \colorbox{gray!30}{\{context\}}\\
\end{tabular} \\ \bottomrule
\end{tabular}
\caption{Prompt template for scenario-based reformulation of legal Q\&A pairs. Part 3 rewrites a statutory Q\&A into a realistic case scenario while preserving the underlying legal reasoning. \colorbox{gray!30}{\{placeholder\}} indicates a slot to be filled with the corresponding value.}
\label{fig:prompt_scenario_reformulation}
\end{figure}
\begin{figure}[ht]
\small
\begin{tabular}{p{\columnwidth}}  
\toprule
\textbf{Part 4. Instruction for Partial Check}\\ \midrule
\begin{tabular}[c]{@{}l@{}}You are given a question describing a legal case.\\ \\ Your task is to determine whether the question can be fully answered using only the given legal provision(s) \\ in context. Please evaluate solely based on the information in the provided context and do not assume any \\legal knowledge beyond it.\\ Use Korean for the justifications.\\ \\ Input:\\ question: \{question\}\\ context: \{context\}\\ \\ Output:\\ Answerable: (Yes / No)\\ Justification: (Explain briefly whether the context alone contains sufficient legal rules or logic to answer the \\question completely.)\\\\ \textbf{<Query>} \\ question: \colorbox{gray!30}{\{question\}}\\ context: \colorbox{gray!30}{\{context\}}
\end{tabular} \\ \midrule
\textbf{Part 5. Instruction for Full Check} \\ \midrule
\begin{tabular}[c]{@{}l@{}}You are given a legal question scenario, its proposed answer, and a set of legal context provisions.\\ \\ Your task is to evaluate the following:\\ \\ 1. Scenario Consistent: Determine whether the background\_scenario + question could have been \\composed using only the explicit legal content given in context. In other words, assess whether the question \\is logically and legally consistent with the context without requiring outside legal knowledge.\\ \\ 2. Correct: Evaluate whether the proposed answer is legally correct based on the background\_scenario, \\ question, and the context.\\ \\ 3. Derivable: Assess whether the proposed answer can be logically and completely derived from the \\ provided context alone, without requiring any unstated assumptions.\\ \\ Use Korean for the justifications.\\ ---\\ Input:\\ background\_scenario: \{background\_scenario\}\\ question: \{question\}\\ answer: \{answer\}\\ context: \{context\}\\ ---\\ Output Format:\\ Scenario Consistent: (Yes / No)\\ Scenario Justification: (Can the question be composed based solely on the given legal context? \\ Explain with reference to the content of the legal provisions.)\\ \\ Correct: (Yes / No)\\ Explanation: (Is the proposed answer legally appropriate in light of the scenario and the legal provisions? \\Provide your reasoning.)\\ \\ Derivable: (Yes / No)\\ Justification: (Can the answer be fully derived from the legal provisions alone? Explain based on \\ the wording and logic of the articles.)\\\\ \textbf{<Query>} \\
background\_scenario: \colorbox{gray!30}{\{background\_scenario\}}\\ question: \colorbox{gray!30}{\{question\}}\\ answer: \colorbox{gray!30}{\{answer\}}\\ context: \colorbox{gray!30}{\{context\}}
\end{tabular} \\ \Xhline{3\arrayrulewidth}
\end{tabular}%
\caption{Prompt templates for validating legal Q\&A pairs. Part 4 checks answerability based on context, while Part 5 evaluates scenario consistency, legal correctness, and derivability. \colorbox{gray!30}{\{placeholder\}} indicates a slot to be filled with the corresponding value.}
\label{fig:prompt_llm_validation}
\end{figure}
\section{Details of \evalName{}}
\label{sec:appendix_LFEval_correlation}
\subsection{Human Evaluation Setup}
\label{sec:appendix_metric_eval_annotator}
To assess the effectiveness and reliability of our proposed metric \evalName, we conduct a human evaluation study comparing it with various metrics including Token-F-1, BLEU \cite{papineni-etal-2002-bleu}, ROUGE-L \cite{lin-2004-rouge} and faithfulness scores \cite{trautmann-etal-2024-measuring}. 

We employ two independent groups of in-house annotators for the evaluation. All samples were anonymized so that annotators were unaware of which model produced each response. All annotators were instructed to consider:
\begin{itemize}
    \item \textbf{Legal relevance}: Whether the response addresses the given question using the provided statutory provisions.
    \item \textbf{Correctness and completeness}: Whether the legal interpretation is accurate and all relevant provisions are applied appropriately.
    \item \textbf{Scoring}: Rate each response on a scale from 1 to 10, where
    \begin{itemize}
        \item 1 indicates a legally irrelevant or unrelated response to the provided context.
        \item 10 indicates a response that makes full and correct use of all provided statutes to reach a legally sound conclusion.
    \end{itemize}
\end{itemize}

\subsection{Correlation with Human Judgment}
\label{sec:appendix_metric_eval_result}

Table~\ref{tab:metric_correlation} reports Pearson correlations between automatic metrics and the averaged human scores.
\textbf{Inter-annotator agreement is high ($r_{agree}=73.97$)}, indicating that the human judgments are consistent and reliable. This level of agreement supports the validity of using averaged human scores as a benchmark for evaluating automatic metrics.

\begin{table}[ht]
\centering
\begin{tabular}{lcc}
\toprule
\textbf{Metric} & \textbf{Pearson $r$} \\
\midrule
\evalName{} (Ours)  & \textbf{84.90} \\
Faithfullness scores \citep{trautmann-etal-2024-measuring} & 6.87  \\
Token-level F-1 & 61.25  \\
BLEU \cite{papineni-etal-2002-bleu} & 51.89\\
ROUGE-L \cite{lin-2004-rouge} & 32.19\\
\bottomrule
\end{tabular}
\caption{Correlation of automatic metrics with human judgments. \evalName{} achieves the strongest alignment with human evaluations.}
\label{tab:metric_correlation}
\end{table}

As shown in the results, \evalName{} exhibits a substantially stronger correlation with human judgments than other standard metrics such as Token F-1 and faithfulness scores. These findings support its effectiveness in capturing the legal fidelity of generated answers in provision-grounded QA tasks.

While the Faithfulness score has shown a strong correlation with human judgment in prior work, it performs poorly when evaluating complex multi-hop legal questions such as those in \datasetName{}.
This indicates that, for complex legal questions, providing detailed and strict evaluation guidelines improves the accuracy and human alignment of LLM-as-a-Judge judgments.

\begin{table}[ht]
\small
\centering
\begin{tabular}{p{0.95\columnwidth}}  
\toprule
\textbf{Examples of \evalName{}} \\ \midrule
\textbf{Question:} Party A and Party B, involved in a civil lawsuit over a construction contract, reached a mutual settlement recorded in the court minutes. Party A later wondered about its legal effect. What is the legal effect of a court-recorded settlement during a civil case? \\ \vspace{0.05em}

\textbf{Context (Gold provisions):} \{Civil Procedure Act Article 220 Effect of Settlement, Waiver of Claim, and Admission of Claim Article 220 (Effect of Settlement, Waiver of Claim, and Admission of Claim) When a settlement, waiver of claim, or admission of claim is recorded in the trial transcript or trial preparation date transcript, the transcript shall have the same effect as a final judgment.\} \\ \vspace{0.05em}

\textbf{Expected output:} same effect as a final judgment. \\ \vspace{0.05em}

\textbf{Generated Answer:} If Party A and Party B reach a settlement in civil litigation in court and the details of the settlement are recorded in the pleadings, the pleadings shall \textbf{have the same effect as a final judgment.} \\ \vspace{0.05em}

\textbf{Score:} \textcolor{green}{10}, \textbf{Reason:} The actual output correctly answers the input question by stating that the record has the same effect as a final judgment, which aligns with the expected output and the legal provision in the context. \\ \midrule

\textbf{Question:} Party A, who runs a small interior design business, traded 70,000 won worth of materials with Party B and wants to apply for transaction confirmation to issue a purchaser-issued invoice. What documents are required, and what conditions must be met for the transaction to be eligible? \\ \vspace{0.05em}

\textbf{Context (Gold provisions):} \{Income Tax Act Enforcement Decree Article 212-4 Issuance of Purchaser-Issued Invoices: Scope and Method, etc. Paragraph 2(2) A resident who intends to issue a purchaser-issued invoice (hereinafter referred to as “purchaser-issued invoice” in this Article) pursuant to Article 163-3(1) of the Act (hereinafter referred to as “purchaser-issued invoice” in this Article) shall issue such invoice within 30 days from the “purchaser-issued invoice”) must submit an application for verification of transaction facts to the tax office director with jurisdiction over the applicant within six months from the end of the tax period in which the supply of goods or services occurred, attaching documents that objectively prove the transaction facts to the transaction fact verification application form prescribed by the Ministry of Economy and Finance. <Amended February 29, 2024>', 'Income Tax Act Enforcement Decree Article 212-4 Issuance of Purchaser-Issued Invoices: Scope and Method, etc. Paragraph 3③ Transactions subject to the transaction fact confirmation application under Paragraph 2 shall be those where the supply price per transaction is 50,000 won or more. <Amended February 29, 2024>\} \\ \vspace{0.05em}

\textbf{Expected output:} Party A must submit an application with supporting documents proving the transaction to the relevant tax office. Only transactions with a supply value of at least 50,000 won are eligible for confirmation. \\ \vspace{0.05em}

\textbf{Generated Answer:} To issue a purchaser-issued invoice, Party A must obtain confirmation from the relevant tax office as the recipient of goods or services. Eligible transactions include cases where the supplier did not issue a tax invoice due to bankruptcy, business closure, or contract cancellation or change. \\ \vspace{0.05em}

\textbf{Score:} \textcolor{red}{2.5344}, \textbf{Reason:} The actual output correctly identifies the need for confirmation from the tax office but omits the requirement to attach objective evidence to the application. It also incorrectly states the conditions for the transaction, missing the (5만원) minimum supply value condition from the context and expected output., \\ 

\bottomrule
\end{tabular}
\caption{\textbf{Examples of \evalName{} Results.} The upper example demonstrates that the generated answer strictly follows both the provision and the expected output, justifying a perfect score of \textcolor{green}{10} with a corresponding explanation. In contrast, the lower example omits critical legal details from the expected output, which is reflected in the reasoning and results in a lower score of \textcolor{red}{2.5344}.}
\label{tab:lf_case}
\end{table}

\subsection{Example of \evalName{} Results}
Unlike conventional metrics that solely rely on gold reference answers and return only a numerical score, \evalName{} offers both the score and an accompanying explanation that justifies the evaluation.
Table~\ref{tab:lf_case} presents evaluation cases on \datasetName{} answers generated by the GPT-4o model using \methodName{}, assessed through \evalName{}.
In addition to providing a fine-grained scalar score via a soft weighted sum based on G-Eval, the evaluation also reveals potential legal issues present in the generated answers.

\section{Experimental Details}
\label{sec:appendix_experiment}
In this section, we provide detailed descriptions of the experimental setup, including the models, evaluation metrics, and baseline methods.

\subsection{Model}
We utilize three LLMs in our experiments: \href{https://huggingface.co/Qwen/Qwen3-32B}{\textit{Qwen/Qwen3-32B}} \cite{qwen3}, \href{https://huggingface.co/LGAI-EXAONE/EXAONE-3.5-32B-Instruct}{\textit{LGAI-EXAONE/EXAONE-3.5-32B-Instruct}} \cite{exaone-3.5}, and \textit{gpt-4o-2024-08-06} \cite{openai2024gpt4technicalreport}.

Qwen is a robust multilingual LLM trained in over 100 languages, including English and Korean, with strong reasoning capabilities.
We found that Qwen3’s powerful reasoning sometimes produced overly long or unfocused outputs, so we ran it in non-thinking mode to keep responses concise and on-target. 
EXAONE is an instruction-tuned English-Korean bilingual LLM with strong performance in Korean QA.
GPT-4o is a high-performing commercial LLM demonstrating state-of-the-art capabilities across diverse tasks and languages.

To ensure reproducibility, we ran the open-source models Qwen3 and EXAONE 3.5 using the vLLM \cite{kwon2023efficient} with nucleus sampling \cite{finlayson2024closing}, with temperature set to 0 and top-p to 0.9. 
These models were served using a single A100 GPU or two A6000 GPUs in tensor parallel mode. 
For GPT-4o, we used the OpenAI API with the same sampling parameters and decoding strategy, incurring a total cost of approximately \$185 for all evaluations and experiments.

\subsection{Evaluation Metrics}
\datasetName\ requires identifying the corresponding legal statutes necessary to resolve complex legal questions. 
Therefore, when evaluating LLM performance on \datasetName, it is essential to assess both retrieval and generation accuracy.

\paragraph{Retrieval.} 
To evaluate retrieval accuracy, we report both Exact Match (EM) and F-1 scores based on provision-level overlap between the predicted and gold statute sets.
Let \( P \) be the set of predicted provisions and \( G \) the set of gold provisions.
The EM score is defined as 1 if \( P = G \) and 0 otherwise.
To compute the F-1 score, we first calculate precision \( p^{\text{prov}} \), recall \( r^{\text{prov}} \), and then the F-1 score \( F_1^{\text{prov}} \), as defined in Equation~\ref{eq:retrieval_f1}.

\begin{equation}
p^{\text{prov}} = \frac{|P \cap G|}{|P|}, \quad
r^{\text{prov}} = \frac{|P \cap G|}{|G|}, \quad
F_1^{\text{prov}} = 
\begin{cases}
\frac{2p^{\text{prov}} r^{\text{prov}}}{p^{\text{prov}} + r^{\text{prov}}} & \text{if } p^{\text{prov}} + r^{\text{prov}} > 0 \\
0 & \text{otherwise}
\end{cases}
\label{eq:retrieval_f1}
\end{equation}

\paragraph{Generation.}
To evaluate the quality of generated answers, we use both token-level F-1 and \evalName\ metrics. 
After normalization and word-level tokenization, the token-level F-1 score is computed by comparing the predicted and ground truth answers. 
Let \( T_{p} \) and \( T_{g} \) denote the sets of word-level tokens from the normalized prediction and ground truth, respectively.
Precision \( p^{\text{token}} \), recall \( r^{\text{token}} \), and token-level F-1 score \( F_{1}^{\text{token}} \) are computed as follows:

\begin{equation}
p^{\text{token}} = \frac{|T_p \cap T_g|}{|T_p|}, \quad
r^{\text{token}} = \frac{|T_p \cap T_g|}{|T_g|}, \quad
F_1^{\text{token}} = 
\begin{cases}
\frac{2p^{\text{token}} r^{\text{token}}}{p^{\text{token}} + r^{\text{token}}} & \text{if } p^{\text{token}} + r^{\text{token}} > 0 \\
0 & \text{otherwise}
\end{cases}
\label{eq:token_f1}
\end{equation}

To compute \evalName, we use GPT-4o with deepeval framework \cite{Ip_deepeval_2025} to generate a score \( s \in [1, 10] \) ten times using the prompt shown in Figure~\ref{fig:eval}. 
For each generated score \( s_i \), we extract its associated token probability \( p(s_i) \) and compute the weighted sum over the 10 generations.
The final \evalName\ score is defined as:

\begin{equation}
\text{\evalName} = \frac{1}{10} \sum_{i=1}^{10} s_i \cdot p(s_i)
\label{eq:lf_eval}
\end{equation}

\subsection{Baselines}
\label{appendix:baseline}
We describe the detailed implementation of our baseline methods. 
Existing research on retrieval-augmented reasoning has primarily focused on general-domain multi-hop QA tasks such as HotpotQA and MuSiQue.
Thus, these approaches do not transfer directly to the legal-domain setting of \datasetName.
Therefore, we re-implemented each baseline by carefully analyzing the original papers and their publicly available code, adapting the logic to suit legal-domain QA better.
We selected the highest-scoring examples among the filtered-out ones and refined them with the help of experts as five in-context learning demonstrations.

\paragraph {Retrieval Module.}
In our experiments, we use BM25 \cite{bm25} retrieval over the statute corpus described in Section \ref{sec:statute_corpus} across all baselines to ensure fairness in retrieval.

\paragraph{Standard Prompting.} 
Standard Prompting (SP) \cite{NEURIPS2020_1457c0d6} directly answers the question using only the model's parametric memory.
We use standard prompting with 5-shot examples. 
As illustrated in Figure \ref{fig:sp_prompt}, we instruct short task descriptions followed by 5-shot examples.
\begin{figure}[ht]
\small
\begin{tabular}{p{\columnwidth}}  
\toprule
\textbf{Standard Prompting prompt} \\ \midrule
You are a legal assistant AI. Given a user Background and a Question, generate a concise and accurate answer. \\
Please provide Answer only without other explanations following the examples.\\ \vspace{0.05em}

\textbf{<Example>} \\
Question: Person A was sentenced to six months in prison in the first instance for assault. Both the defendant and the prosecutor appealed. 
Can the appellate court sentence Person A to a longer sentence than the first trial? \\ 
Answer: The appellate court cannot sentence Person A to a longer sentence than the first trial.\\ \vspace{0.05em}
 \{Example\} (Continue to 5 examples)\\ \vspace{0.05em}
\textbf{<Query>} \\
Question: \colorbox{gray!30}{\{background\}+\{question\}} \\
Answer: \\
\bottomrule
\end{tabular}
\caption{Prompt for \textbf{Standard Prompting (SP)}.
We employ 5 different examples.
\colorbox{gray!30}{\{placeholder\}} indicates a slot to be filled with the corresponding value for evaluation.}
\label{fig:sp_prompt}
\end{figure}

\paragraph{Chain of Thought.}
Chain of Thought (CoT) \cite{NEURIPS2022_9d560961} generates reasoning steps relying on the parametric memory before the final answer.
We use Chain-of-Thought prompting with 5-shot examples.
As illustrated in Figure \ref{fig:cot_prompt}, we instruct short task descriptions followed by 5-shot examples.
The reasoning traces were initially generated by GPT-4o and subsequently refined by human annotators.
\begin{figure}[ht]
\small
\begin{tabular}{p{\columnwidth}} 
\toprule
\textbf{Chain-of-Thought prompt} \\ \midrule
You are a helpful assistant. Whenever you are given a legal (or similar) question, structure your response into exactly two sections: \\
1. “Reasoning:”  \\
– List each logical step you took to arrive at the conclusion.  \\
– Use numbered or bulleted items, each beginning with a brief statement of fact or rule. \\
– Make the chain of thought clear and concise. 
\\ 
2. “Answer:”  \\
– Provide a single, direct, definitive answer to the question in one or two sentences. \\
– Do not include any additional reasoning here—just the conclusion. \\ \vspace{0.05em}

\textbf{<Example>} \\
Question: Person A was sentenced to six months in prison in the first instance for assault. Both the defendant and the prosecutor appealed. 
Can the appellate court sentence Person A to a longer sentence than the first trial? \\ 
Reasoning: In the first trial, Person A was sentenced to 6 months in prison for assault, and both the defendant and the prosecutor appealed the judgment. The scope of appellate review is limited to the “reasons for appeal” (Criminal Code §372), and the judgment can only be changed within the reasons given by the defendant and the prosecutor. Criminal Code §364(2) states that “the sentence of the final judgment may be reduced, exempted, or changed, but not aggravated, when the defendant appeals.” The non-commutation principle applies. Because the prosecutor did not request an “aggravated sentence” as a ground for appeal, the appellate court cannot consider a sentence that is more severe than the first sentence. Therefore, the appellate court cannot sentence Person A to a sentence that is more severe than the first sentence. \\
Answer: The appellate court cannot sentence Person A to a longer sentence than the first trial.\\ \vspace{0.05em}
 \{Example\} (Continue to 5 examples)\\ \vspace{0.05em}
\textbf{<Query>} \\
Question: \colorbox{gray!30}{\{background\}+\{question\}} \\
Reasoning: \\
\bottomrule
\end{tabular}
\caption{Prompt for \textbf{Chain-of-Thought (CoT)}.
We employ 5 different examples.
\colorbox{gray!30}{\{placeholder\}} indicates a slot to be filled with the corresponding value for evaluation.}
\label{fig:cot_prompt}
\end{figure}

\paragraph{One-time Retrieval.}
One-time Retrieval (OR) uses the original question to retrieve and augment the top-$n$ provisions, where $n$ is the number of reasoning hops in each \datasetName{} instance.
We employ a BM-25 retriever for retrieving provisions. We implement retrieval module via open source toolkit \cite{bm25s}. 
In the case of one-time retrieval, we provide the top-n provisions as retrieved-context in Figure~\ref{fig:or_prompt}, where n corresponds to the number of reasoning hops the question requires.
\begin{figure}[ht]
\small
\begin{tabular}{p{\columnwidth}}  
\toprule
\textbf{Retrieval Augmented prompt} \\ \midrule
\colorbox{gray!30}{\{Standard Prompting System Prompt\} or \{Chain-of-Thought System Prompt\}} \\ \vspace{0.05em}

\textbf{<Example>} \\
Context: Article 368 of the Code of Criminal Procedure Article 368 (Prohibition of adverse changes) “In cases appealed by the accused and in cases appealed on behalf of the accused, the court shall not impose a sentence heavier than that of the original judgment.”\\
Question: Person A was sentenced to six months in prison in the first instance for assault. Both the defendant and the prosecutor appealed. \\ 
Optional: \{Reasoning: In the first trial, Person A was sentenced to 6 months in prison for assault, and both the defendant and the prosecutor appealed the judgment. The scope of appellate review is limited to the “reasons for appeal” (Criminal Code §372), and the judgment can only be changed within the reasons given by the defendant and the prosecutor. Criminal Code §364(2) states that “the sentence of the final judgment may be reduced, exempted, or changed, but not aggravated, when the defendant appeals.” The non-commutation principle applies. Because the prosecutor did not request an “aggravated sentence” as a ground for appeal, the appellate court cannot consider a sentence that is more severe than the first sentence. Therefore, the appellate court cannot sentence Person A to a sentence that is more severe than the first sentence.\} \\
Answer: The appellate court cannot sentence Person A to a longer sentence than the first trial.\\ \vspace{0.05em}
 \{Example\} (Continue to 5 examples)\\ \vspace{0.05em}
\textbf{<Query>} \\
Question: \colorbox{gray!30}{\{background\}+\{question\}} \\
Context: \colorbox{gray!30}{\{Retrieved Contexts\}}\\
Answer: \\
\bottomrule
\end{tabular}
\caption{Prompt for retrieval augmented question answering.
The system prompt and inclusion of reasoning traces vary depending on whether Self-Prediction (SP) or Chain-of-Thought (CoT) prompting is used. The retrieval method \methodName{} adopts the SP setting.
We employ 5 different examples.
\colorbox{gray!30}{\{placeholder\}} indicates a slot to be filled with the corresponding value for evaluation.}
\label{fig:or_prompt}
\end{figure}

\paragraph{Self-Ask.}
Self-Ask \cite{press-etal-2023-measuring} iteratively determines whether follow-up questions are needed and generates intermediate queries.
It then retrieves reference provisions and uses the augmented context to produce the final answer.
We implement Self-Ask following the officially implemented code.\footnote{\href{https://github.com/ofirpress/self-ask}{ofirpress/self-ask}}
As the original implementation is designed for the general domain using the Google search engine, we adapt the retrieval module to use our own retriever instead. 
We set a termination condition by limiting the reasoning depth up to 5 for GPT-4o and 8 for Qwen and EXAONE.
We employ the prompt in Figure \ref{fig:selfask_prompt}.

\begin{figure}[ht]
\small
\begin{tabular}{p{\columnwidth}}  
\toprule
\textbf{Self-Ask prompt} \\ \midrule
You are a self-ask legal reasoning assistant. When given a new Question, follow this format exactly, with no deviations:\\ 
Question: <question> \\
Are follow up questions needed here: <Yes or No>\\
If Yes:\\
Follow up: <one specific clarifying question>\\
Intermediate answer: <brief grounded answer>\\
(repeat Follow up and Intermediate answer pairs until you have all facts)\\
So the final answer is: <your internal reasoning summary>\\
\vspace{0.05em}

\textbf{<Example>} \\
Question: Person A was sentenced to six months in prison in the first instance for assault. Both the defendant and the prosecutor appealed. 
Can the appellate court sentence Person A to a longer sentence than the first trial? \\ 
Are follow up question needed here: Yes \\
Follow up: In what cases can a sentence be more severe on appeal?
Intermediate answer: Because the prosecutor did not request an “aggravated sentence” as a ground for appeal, the appellate court cannot consider a sentence that is heavier than the first trial. \\
So the final answer is: The appellate court cannot sentence Person A to a longer sentence than the first trial.\\ \vspace{0.05em}
 \{Example\} (Continue to 5 examples)\\ \vspace{0.05em}
\textbf{<Query>} \\
Question: \colorbox{gray!30}{\{background\}+\{question\}} \\
Are follow up question needed here: \\
\bottomrule
\end{tabular}
\caption{Prompt for \textbf{Self-ask}.
We employ 5 different examples.
\colorbox{gray!30}{\{placeholder\}} indicates a slot to be filled with the corresponding value for evaluation.}
\label{fig:selfask_prompt}
\end{figure}
\paragraph{IRCOT.}
IRCoT \cite{trivedi-etal-2023-interleaving} interleaves CoT traces with retrieval, using the interleaved generations to incorporate external knowledge into the reasoning process.
We implement IRCoT following the officially implemented code.\footnote{\href{https://github.com/StonyBrookNLP/ircot}{StonyBrookNLP/ircot}}
As the original implementation is designed for the general domain using the Elastic search engine, we adapt the retrieval module to use our own retriever instead.
We set a termination condition by limiting the reasoning depth up to 5 for GPT-4o and 8 for Qwen and EXAONE.
We employ the prompt in Figure \ref{fig:ircot_flare_prompt}.

\begin{figure}[ht]
\small
\begin{tabular}{p{\columnwidth}}  
\toprule
\textbf{IRCoT \& FLARE prompt} \\ \midrule
You are a self-ask legal reasoning assistant. When given a context and a question, output exactly the following plain-text template.\\ 
Context: <legal provisions> \\
Question: <question>\\
Answer:<step-by-step reasons for the final answer> So the final answer is: <final answer>\\
\vspace{0.05em}

\textbf{<Example>} \\
Context: Article 368 of the Code of Criminal Procedure Article 368 (Prohibition of adverse changes) “In cases appealed by the accused and in cases appealed on behalf of the accused, the court shall not impose a sentence heavier than that of the original judgment.”\\
Question: Person A was sentenced to six months in prison in the first instance for assault. Both the defendant and the prosecutor appealed. 
Can the appellate court sentence Person A to a longer sentence than the first trial? \\ \vspace{0.05em}
Answer: In the first trial, Person A was sentenced to 6 months in prison for assault, and both the defendant and the prosecutor appealed the judgment. The scope of appellate review is limited to the “reasons for appeal” (Criminal Code §372), and the judgment can only be changed within the reasons given by the defendant and the prosecutor. Criminal Code §364(2) states that “the sentence of the final judgment may be reduced, exempted, or changed, but not aggravated, when the defendant appeals.” The non-commutation principle applies. Because the prosecutor did not request an “aggravated sentence” as a ground for appeal, the appellate court cannot consider a sentence that is more severe than the first sentence. \\
So the final answer is: The appellate court cannot sentence Person A to a sentence that is more severe than the first sentence. \\ \vspace{0.05em}
 \{Example\} (Continue to 5 examples)\\ \vspace{0.05em}
\textbf{<Query>} \\
Question: \colorbox{gray!30}{\{background\}+\{question\}} \\
Answer: \\ \midrule

\textbf{FLARE query generation prompt} \\ \midrule
The following user query has been partially masked due to low-confidence tokens.\\
Please review the masked query and formulate a Korean question that would allow you to search for the most relevant legal provisions needed to answer the question.\\ \vspace{0.05em}

\colorbox{gray!30}{\{question\}} \\
Query: \colorbox{gray!30}{\{query\}}\\
New Query: \\

\bottomrule
\end{tabular}
\caption{Prompt for \textbf{IRCoT and FLARE}.
We employ 5 different examples.
\colorbox{gray!30}{\{placeholder\}} indicates a slot to be filled with the corresponding value for evaluation.}
\label{fig:ircot_flare_prompt}
\end{figure}
\paragraph{FLARE.}
FLARE \cite{jiang-etal-2023-active} interleaves reasoning and retrieval similar to IRCoT but selectively performs retrieval only for reasoning steps with low confidence.
We implement IRCoT following the officially implemented code.\footnote{\href{https://github.com/jzbjyb/FLARE}{jzbjyb/FLARE}}
As the original implementation is designed for the general domain using the Bing search engine, we adapt the retrieval module to use our own retriever instead.
Due to several configurable parameters in FLARE, we set the log-probability threshold to -1.5 for Qwen and EXAONE and -0.6 for GPT-4o to achieve a retrieval rate close to 50\%, which was reported as optimal in the reference paper. 
We adapt the instruct mode to generate retrieval queries.
We set a termination condition by limiting the reasoning depth up to 5 for GPT-4o and 8 for Qwen and EXAONE.
The prompt is in the Figure \ref{fig:ircot_flare_prompt}.

\paragraph{ProbTree.}
\begin{figure}[ht]
\small
\begin{tabular}{p{\columnwidth}}  
\toprule
\textbf{Closebook prompt} \\ \midrule
You are given legal Q\&A examples. For a new legal question, answer briefly and clearly in one or two sentences.\\ \vspace{0.05em}

\textbf{<Example>} \\
Question: Can the appellate court sentence Person A to a longer sentence than the first trial? \\ 
Answer: The appellate court cannot sentence Person A to a longer sentence than the first trial.\\ \vspace{0.05em}
 \{Example\} (Continue to 5 examples)\\ \vspace{0.05em}
\textbf{<Query>} \\
Question: \colorbox{gray!30}{\{question\}} \\
Answer: \\ \midrule
\textbf{Openbook prompt} \\ \midrule
You are given a legal question and its related law texts (context). Read the context carefully and write a concise, plain-text answer (1–2 sentences) that accurately summarizes the legal principle or outcome.\\ \vspace{0.05em}

\textbf{<Example>} \\
Question: Can the appellate court sentence Person A to a longer sentence than the first trial? \\ 
Context: Article 368 of the Code of Criminal Procedure Article 368 (Prohibition of adverse changes) “In cases appealed by the accused and in cases appealed on behalf of the accused, the court shall not impose a sentence heavier than that of the original judgment.”\\
Answer: The appellate court cannot sentence Person A to a longer sentence than the first trial.\\ \vspace{0.05em}
 \{Example\} (Continue to 5 examples)\\ \vspace{0.05em}
\textbf{<Query>} \\
Question: \colorbox{gray!30}{\{question\}} \\
Answer: \\ \midrule
\textbf{Child Aggregate prompt} \\ \midrule
You are given a context and a legal question. \\
Use only the information from the provided context to write a concise and accurate legal answer to the question. \\ \vspace{0.5em}
\textbf{<Example>} \\
Context: Can the appellate court sentence Person A to a longer sentence than the first trial? The appellate court cannot sentence Person A to a longer sentence than the first trial. \\
Question: Person A was sentenced to six months in prison in the first instance for assault. Both the defendant and the prosecutor appealed. 
Can the appellate court sentence Person A to a longer sentence than the first trial? \\ 
Answer: In the first trial, Person A was sentenced to 6 months in prison for assault, and both the defendant and the prosecutor appealed the judgment. The scope of appellate review is limited to the “reasons for appeal” (Criminal Code §372), and the judgment can only be changed within the reasons given by the defendant and the prosecutor. Criminal Code §364(2) states that “the sentence of the final judgment may be reduced, exempted, or changed, but not aggravated, when the defendant appeals.” The non-commutation principle applies. Because the prosecutor did not request an “aggravated sentence” as a ground for appeal, the appellate court cannot consider a sentence that is more severe than the first sentence. Therefore, the appellate court cannot sentence Person A to a sentence that is more severe than the first sentence. \\ \vspace{0.5em}
 \{Example\} (Continue to 5 examples)\\ \vspace{0.05em}
\textbf{<Query>} \\
Question: \colorbox{gray!30}{\{background\}+\{question\}} \\
Answer: \\
\bottomrule
\end{tabular}
\caption{Prompt for \textbf{ProbTree and BeamAggr}.
We employ 5 different examples.
\colorbox{gray!30}{\{placeholder\}} indicates a slot to be filled with the corresponding value for evaluation.}
\label{fig:probtree}
\end{figure}
ProbTree  \cite{cao-etal-2023-probabilistic} decomposes the question into a tree structure and explores multiple reasoning strategies at each node. Final answers are selected by aggregating candidates based on their generation log probabilities.
We implement ProbTree following the officially implemented code.\footnote{\href{https://github.com/THU-KEG/ProbTree}{THU-KEG/ProbTree}}
As the original implementation is designed for the general domain using the Elastic search engine, we adapt the retrieval module to use our own retriever instead.
For tree generation, we followed the prompt in the official implementation.
We only include the retrieved context from the retrieval module when the $\{openbook\}$ is selected.
Prompt used for $\{closebook, openbook, child\ aggregation\}$ is in Figure \ref{fig:probtree}.

\paragraph{BeamAggr.}
BeamAggr \cite{chu-etal-2024-beamaggr} enhances ProbTree with multi-source reasoning and probabilistic answer aggregation. 
Since the official implementation code is unavailable, we implement BeamAggr from scratch by closely following the descriptions in the reference paper.
Unlike the experiments conducted in the original paper, \datasetName{} focuses on the legal domain. Accordingly, we adapt the retrieval module and restrict multi-source reasoning to ${ \text{closebook}, \text{openbook} }$ in order to obtain the distribution over each leaf node.
Furthermore, due to the lack of sufficient sources, we utilize the log probabilities of generated tokens to estimate the response probabilities from each source.
For tree generation, we followed the prompt in the official implementation of ProbTree.
We only include the retrieved context from the retrieval module when the $\{openbook\}$ is selected.
The prompt for $\{closebook, openbook\}$ is in Figure \ref{fig:probtree}.

\paragraph{\methodName{}.}
\begin{figure}[ht]
\small
\begin{tabular}{p{\columnwidth}}  
\toprule
\textbf{Parametric provision generation prompt} \\ \midrule
You are an expert legal assistant whose task is to identify and return all relevant statutory provisions that support the answer to a given legal question.\\
Your role is not to provide interpretations, summaries, or conclusions, but to retrieve and list the exact legal clauses that serve as a legal basis for the scenario described.\\ 
- Answer must be a list of clauses in the following format: ["Name of Law (Title or Clause Summary) Exact clause text or its key portion.","Name of Law (Title or Clause Summary) Exact clause text or its key portion.",...] without any other explanations. \\
- If multiple laws are involved, list all clauses together in a single list.\\ 
- If no directly applicable statutory provision exists, generate the most plausible clause in the same format, as if it were part of the relevant law. \\ \vspace{0.05em}

\textbf{<Example>} \\
Question: Person A was sentenced to six months in prison in the first instance for assault. Both the defendant and the prosecutor appealed. 
Can the appellate court sentence Person A to a longer sentence than the first trial? \\ 
Answer: [ “Article of the Code of Criminal Procedure Article  (Prohibition of adverse changes) In cases appealed by the accused and in cases appealed on behalf of the accused, the court shall not impose a sentence heavier than that of the original judgment.”]\\ \vspace{0.05em}
 \{Example\} (Continue to 5 examples)\\ \vspace{0.05em}
\textbf{<Query>} \\
Question: \colorbox{gray!30}{\{background\}+\{question\}} \\
Answer: \\ \midrule
\textbf{Selection prompt} \\ \midrule
You are given a question, and a list of candidate passages with associated passage IDs.  
Your task is to identify the most proper passage that directly support the answer to the question.  
Please select only one ID among given candidate passages. \\
Even if multiple passages seem relevant or none seem perfectly appropriate, you must select exactly one passage ID. \\ \vspace{0.5em}

Background: Person A was sentenced to six months in prison in the first instance for assault. Both the defendant and the prosecutor appealed. \\
Question: Can the appellate court sentence Person A to a longer sentence than the first trial?\\
Candidates:\\
0: Article of the Code of Criminal Procedure Article  (Prohibition of adverse changes) In cases appealed by the accused and in cases appealed on behalf of the accused, the court shall not impose a sentence heavier than that of the original judgment. \\
1: Article 274 of the Code of Military Justice, Cancellation of Charges, Paragraph 3 ③ Paragraphs 1 and 2 shall also apply to the withdrawal of an expression of desire for punishment in cases where the victim cannot be charged against expressed will.\\
...  \\ (Top-10 provisions are placed here)\\ 
Answer: 0 \\ \vspace{0.05em}
 \{Example\} (Continue to 5 examples)\\ \vspace{0.05em}
 \textbf{<Query>} \\
Background: \colorbox{gray!30}{\{background\}} \\
Question: \colorbox{gray!30}{\{question\}} \\
Context: \colorbox{gray!30}{\{idx\}: \{provision\}} (Aggregate Top-10 provisions) \\
Answer: \\ 
\bottomrule
\end{tabular}
\caption{Prompt for \textbf{\methodName}.
For parametric provision generation, we instruct LLM to generate list of parametric provisions.
For selection, we instruct LLM to select most relevant provision among top-10 candidates.
We employ 5 different examples.
\colorbox{gray!30}{\{placeholder\}} indicates a slot to be filled with the corresponding value for evaluation.}
\label{fig:parser_prompt}
\end{figure}
For parametric provision generation, we instruct the LLM to generate list-style text and extract the results using a JSON parser.
For the Retrieve, Rerank, and Selection retrieval pipeline, we utilize three different language models, one for each stage.
In the Retrieve stage, our retrieval module retrieves the top 100 provisions most relevant to the generated parametric provision from the statute corpus.
In the Rerank stage, we use a fine-tuned BGE reranker\footnote{\href{https://huggingface.co/dragonkue/bge-reranker-v2-m3-ko}{dragonkue/bge-reranker-v2-m3-ko}}\cite{chen2024bge} to sort the top 100 retrieved provisions based on relevance.
In the Selection stage, an LLM generates the ID of the most relevant provision among the top 10. This selected provision is then used as the supporting legal provision.

\section{Case Study}
To better understand how existing baseline methods perform on open-ended and multi-hop legal questions, we conduct a case study on \datasetName. Figure~\ref{fig:casestudy1} and \ref{fig:casestudy2} show English-translated responses from various baseline methods, while Figure \ref{fig:casestudy3} and \ref{fig:casestudy4} illustrate Korean responses from various baseline methods.
Most baseline methods fail to reason over multiple relevant provisions.
By contrast, \methodName{} successfully reasons over all three reference statutes and produces a coherent, legally reliable answer. 
\begin{figure*}[ht]
\small
\begin{tabular}{p{\linewidth}} 
\toprule
\textbf{Background Scenario} \\ \midrule
Person A is the head of a newly established public opinion polling agency, "Trend Survey," and plans to conduct a public opinion poll related to the upcoming local elections and subsequently publish the results in the media. To this end, Person A has equipped the agency with a survey system and analysis specialists, and has fulfilled the requirements set forth by the Central Election Management Committee's regulations. Person A has submitted a written application for registration with the competent Election Survey Deliberation Commission. Person A is aware that once the registration process is completed, information about the agency will be made available to the general public.\\ \midrule
\textbf{Question} \\ \midrule
 What requirements and procedures must "Trend Survey" follow to conduct and publish or report a public opinion poll related to elections? After Person A submits the registration application, what procedures must the competent Election Survey Deliberation Commission follow? Additionally, how is information about "Trend Survey" handled after the issuance of the registration certificate?\\ \midrule
\textbf{Gold Answer} \\ \midrule
"Trend Survey" must meet \textcolor{red}{the requirements set by the Central Election Management Committee's regulations, such as having a survey system and analysis specialists, and must apply for registration in writing with the competent Election Survey Deliberation Commission.} \textcolor{blue}{The competent Election Survey Deliberation Commission must process the registration within 7 days from the date of receipt of the application and issue a registration certificate.} Once the registration certificate is issued, \textcolor{orange}{the Election Survey Deliberation Commission must promptly disclose information about the agency on the Central Election Survey Deliberation Commission's website}, in accordance with the regulations set by the Central Election Management Committee.\\ \midrule
\textbf{Gold supporting provisions} \\ \midrule
\textcolor{red}{\textbf{PUBLIC OFFICIAL ELECTION ACT / Article. 8-9 / Registration of Polling Institutions or Organizations / Paragraph. 1}} \\
Where a polling institution or organization intends to conduct public opinion polls relating to election for the purpose of publishing or reporting, it shall file an application for the registration of the polling institution or organization with the competent deliberative committee on public opinion polls relating to election in writing after it is equipped with an examination system and experts in analysis and meets other requirements prescribed by the National Election Commission Regulations.\\
\textcolor{blue}{\textbf{PUBLIC OFFICIAL ELECTION ACT / Article. 8-9 / Registration of Polling Institutions or Organizations / Paragraph. 2}}\\
The competent deliberative committee on public opinion polls relating to election upon receiving an application for registration under paragraph (1) shall accept the registration and issue a certificate of registration within seven days from the date it receives such application.\\
\textcolor{orange}{\textbf{PUBLIC OFFICIAL ELECTION ACT / Article. 8-9 / Registration of Polling Institutions or Organizations / Paragraph. 3}} \\
The competent deliberative committee on public opinion polls relating to election shall without delay disclose information prescribed by the National Election Commission Regulations, which is information on the polling institution or organization (hereinafter referred to as "institution conducting public opinion polls relating to election") to which it issues a certificate of registration pursuant to paragraph (2), on the website of the National Election Survey Deliberation Commission.
\\ \midrule
\textbf{\methodName{} (Ours)} \\
If "Trend Survey" intends to conduct public opinion polls related to elections for the purpose of publication or reporting, \textcolor{red}{it must be equipped with a survey system, professional analysis personnel, and meet the requirements set forth by the rules of the National Election Commission, then submit a written registration application to the competent Election Poll Deliberation Commission.} After the application is submitted by "Trend Survey," \textcolor{blue}{the competent commission must process the registration and issue a registration certificate within seven days from the date of receipt.} Once the certificate is issued, \textcolor{orange}{information about "Trend Survey," as specified by the rules of the National Election Commission, must be promptly disclosed on the website of the Central Election Poll Deliberation Commission.}\\ 
\textbf{Standard Prompting (SP)} \\
If "Trend Survey" intends to conduct public opinion polls related to elections for the purpose of publication or reporting, it must meet the requirements set forth by the rules of the National Election Commission and register with the competent Election Poll Deliberation Commission. After the registration application is submitted, the commission reviews whether the requirements are met and then issues a registration certificate. Once the certificate is issued, information regarding "Trend Survey" is disclosed to the general public. \\ 
\textbf{Chain-of-Thought (CoT)} \\
If "Trend Survey" intends to conduct public opinion polls related to elections for the purpose of publication or reporting, it must meet the requirements prescribed by the rules of the National Election Commission and register with the competent Election Poll Deliberation Commission. Once the registration certificate is issued, information is disclosed to the general public.\\ 
\textbf{Self-Ask} \\ 
Are follow up questions needed here: Yes. (Continue) \\ 
\textbf{IRCoT} \\
"Trend Survey" \textcolor{red}{must be equipped with a survey system and professional analysis personnel, and after meeting the requirements specified by the rules of the National Election Commission, it must submit a written registration application to the competent Election Poll Deliberation Commission.} The competent commission \textcolor{blue}{must process the registration and issue a registration certificate within seven days from the date of receiving the application.}\\
\bottomrule
\end{tabular}
\caption{Case study of a single English instance from \datasetName{}. The contents of each gold supporting provision are highlighted in different colors. While both SP and CoT capture the overall reasoning flow, they fail to identify the specific contents of the provisions. Self-Ask fail to answer the question. In IRCoT, one of the gold provisions is missing.}
\label{fig:casestudy1}
\end{figure*}

\begin{figure*}[ht]
\small
\begin{tabular}{p{\linewidth}}  
\toprule
\textbf{Background Scenario} \\ \midrule
Person A is the head of a newly established public opinion polling agency, "Trend Survey," and plans to conduct a public opinion poll related to the upcoming local elections and subsequently publish the results in the media. To this end, Person A has equipped the agency with a survey system and analysis specialists, and has fulfilled the requirements set forth by the Central Election Management Committee's regulations. Person A has submitted a written application for registration with the competent Election Survey Deliberation Commission. Person A is aware that once the registration process is completed, information about the agency will be made available to the general public.\\ \midrule
\textbf{Question} \\ \midrule
 What requirements and procedures must "Trend Survey" follow to conduct and publish or report a public opinion poll related to elections? After Person A submits the registration application, what procedures must the competent Election Survey Deliberation Commission follow? Additionally, how is information about "Trend Survey" handled after the issuance of the registration certificate?\\ \midrule
\textbf{Gold Answer} \\ \midrule
"Trend Survey" must meet \textcolor{red}{the requirements set by the Central Election Management Committee's regulations, such as having a survey system and analysis specialists, and must apply for registration in writing with the competent Election Survey Deliberation Commission.} \textcolor{blue}{The competent Election Survey Deliberation Commission must process the registration within 7 days from the date of receipt of the application and issue a registration certificate.} Once the registration certificate is issued, \textcolor{orange}{the Election Survey Deliberation Commission must promptly disclose information about the agency on the Central Election Survey Deliberation Commission's website}, in accordance with the regulations set by the Central Election Management Committee.\\ \midrule
\textbf{Gold supporting provisions} \\ \midrule
\textcolor{red}{\textbf{PUBLIC OFFICIAL ELECTION ACT / Article. 8-9 / Registration of Polling Institutions or Organizations / Paragraph. 1}} \\
Where a polling institution or organization intends to conduct public opinion polls relating to election for the purpose of publishing or reporting, it shall file an application for the registration of the polling institution or organization with the competent deliberative committee on public opinion polls relating to election in writing after it is equipped with an examination system and experts in analysis and meets other requirements prescribed by the National Election Commission Regulations.\\
\textcolor{blue}{\textbf{PUBLIC OFFICIAL ELECTION ACT / Article. 8-9 / Registration of Polling Institutions or Organizations / Paragraph. 2}}\\
The competent deliberative committee on public opinion polls relating to election upon receiving an application for registration under paragraph (1) shall accept the registration and issue a certificate of registration within seven days from the date it receives such application.\\
\textcolor{orange}{\textbf{PUBLIC OFFICIAL ELECTION ACT / Article. 8-9 / Registration of Polling Institutions or Organizations / Paragraph. 3}} \\
The competent deliberative committee on public opinion polls relating to election shall without delay disclose information prescribed by the National Election Commission Regulations, which is information on the polling institution or organization (hereinafter referred to as "institution conducting public opinion polls relating to election") to which it issues a certificate of registration pursuant to paragraph (2), on the website of the National Election Survey Deliberation Commission.
\\ \midrule
\textbf{\methodName{} (Ours)} \\
If "Trend Survey" intends to conduct public opinion polls related to elections for the purpose of publication or reporting, \textcolor{red}{it must be equipped with a survey system, professional analysis personnel, and meet the requirements set forth by the rules of the National Election Commission, then submit a written registration application to the competent Election Poll Deliberation Commission.} After the application is submitted by "Trend Survey," \textcolor{blue}{the competent commission must process the registration and issue a registration certificate within seven days from the date of receipt.} Once the certificate is issued, \textcolor{orange}{information about "Trend Survey," as specified by the rules of the National Election Commission, must be promptly disclosed on the website of the Central Election Poll Deliberation Commission.}\\ 
\textbf{FLARE} \\
"Trend Survey" \textcolor{red}{must be equipped with a survey system and professional analysis personnel, and after meeting the requirements specified by the rules of the National Election Commission, it must submit a written registration application to the competent Election Poll Deliberation Commission.} After the application is submitted, the commission reviews whether the requirements are met and issues a registration certificate. \\ 
\textbf{ProbTree} \\
If "Trend Survey" intends to conduct public opinion polls related to elections for the purpose of publication or reporting, \textcolor{red}{it must meet the requirements set by the rules of the National Election Commission—such as having a survey system and professional analysis personnel—and submit a written registration application to the competent Election Poll Deliberation Commission.} After the application is submitted, the commission reviews the registration in accordance with relevant procedures, and once the registration certificate is issued, information about "Trend Survey" is managed accordingly.\\ 
\textbf{BeamAGgr} \\
To conduct public opinion polls related to elections for the purpose of publication or reporting, a polling agency must meet the requirements set by the rules of the National Election Commission and submit a written registration application to the competent Election Poll Deliberation Commission. The use of virtual mobile phone numbers is permitted.\\
\bottomrule
\end{tabular}
\caption{Case study of a single English instance from \datasetName{}. The contents of each gold supporting provision are highlighted in different colors. FLARE and ProbTree correctly refer to one provision but miss the other two. BeamAggr provides a vague explanation and includes a hallucinated answer.}
\label{fig:casestudy2}
\end{figure*}

\begin{figure*}[ht]
\small
\begin{tabular}{p{\linewidth}}  
\toprule
\textbf{Background Scenario} \\ \midrule
갑은 신생 여론조사기관 '트렌드서베이'의 대표로, 다가오는 지방선거와 관련된 여론조사를 실시한 후 그 결과를 언론에 공표하려는 계획을 가지고 있다. 이를 위해 갑은 조사 시스템과 분석 전문 인력을 갖추고, 중앙선거관리위원회규칙에서 요구하는 요건을 충족한 뒤 관할 선거여론조사심의위원회에 등록을 서면으로 신청하였다. 등록 절차가 완료되면 해당 기관에 관한 정보가 일반 국민에게도 공개된다는 점을 인지하고 있다.\\ \midrule
\textbf{Question} \\ \midrule
'트렌드서베이'가 선거에 관한 여론조사를 공표하거나 보도 목적으로 실시하려면 어떤 요건과 절차를 거쳐야 하며, 갑의 등록 신청 이후 관할 선거여론조사심의위원회는 어떤 처리 절차를 따라야 하는가? 또한, 등록증이 교부된 이후 '트렌드서베이'에 관한 정보는 어떻게 처리되는가?\\ \midrule
\textbf{Gold Answer} \\ \midrule
'트렌드서베이'는 \textcolor{red}{조사 시스템, 분석 전문 인력 등 중앙선거관리위원회규칙으로 정한 요건을 갖추고 관할 선거여론조사심의위원회에 서면으로 등록을 신청해야 한다.} 관할 선거여론조사심의위원회는 \textcolor{blue}{등록신청을 접수한 날부터 7일 이내에 등록을 수리하고 등록증을 교부해야 한다.} 등록증이 교부되면, 선거여론조사심의위원회는 해당 기관에 관한 정보를 중앙선거관리위원회규칙으로 정하는 바에 따라 \textcolor{orange}{지체 없이 중앙선거여론조사심의위원회 홈페이지에 공개해야 한다.}\\ \midrule
\textbf{Gold supporting provisions} \\ \midrule
\textcolor{red}{공직선거법 8조 9 여론조사 기관ㆍ단체의 등록 등 1항} \\
① 여론조사 기관ㆍ단체가 공표 또는 보도를 목적으로 선거에 관한 여론조사를 실시하려는 때에는 조사시스템, 분석전문인력, 그 밖에 중앙선거관리위원회규칙으로 정하는 요건을 갖추어 관할 선거여론조사심의위원회에 서면으로 그 등록을 신청하여야 한다.\\
\textcolor{blue}{공직선거법 8조 9 여론조사 기관ㆍ단체의 등록 등 2항}\\
② 제1항에 따른 등록신청을 받은 관할 선거여론조사심의위원회는 그 신청을 접수한 날부터 7일 이내에 등록을 수리하고 등록증을 교부하여야 한다.\\
\textcolor{orange}{공직선거법 8조 9 여론조사 기관ㆍ단체의 등록 등 3항} \\
③ 선거여론조사심의위원회는 제2항에 따라 등록증을 교부한 여론조사 기관ㆍ단체(이하 "선거여론조사기관"이라 한다)에 관한 정보로서 중앙선거관리위원회규칙으로 정하는 정보를 지체 없이 중앙선거여론조사심의위원회 홈페이지에 공개하여야 한다.
\\ \midrule
\textbf{\methodName{} (Ours)} \\
'트렌드서베이'가 선거에 관한 여론조사를 공표하거나 보도 목적으로 실시하려면 \textcolor{red}{조사 시스템, 분석 전문 인력, 중앙선거관리위원회규칙으로 정하는 요건을 갖추어 관할 선거여론조사심의위원회에 서면으로 등록을 신청해야 한다.} 갑의 등록 신청 이후, 관할 선거여론조사심의위원회는 \textcolor{blue}{신청을 접수한 날부터 7일 이내에 등록을 수리하고 등록증을 교부해야 한다.} 등록증이 교부된 이후, '트렌드서베이'에 관한 정보는 중앙선거관리위원회규칙으로 정하는 정보를 \textcolor{orange}{지체 없이 중앙선거여론조사심의위원회 홈페이지에 공개해야 한다.}\\ \vspace{0.5em}
\textbf{Standard Prompting (SP)} \\
'트렌드서베이'가 선거에 관한 여론조사를 공표하거나 보도 목적으로 실시하려면 중앙선거관리위원회규칙에서 정한 요건을 충족하고 관할 선거여론조사심의위원회에 등록해야 하며, 등록 신청 후 위원회는 요건 충족 여부를 심사하여 등록증을 교부한다. 등록증이 교부된 이후 '트렌드서베이'에 관한 정보는 일반 국민에게 공개된다. \\ \vspace{0.5em}
\textbf{Chain-of-Thought (CoT)} \\
'트렌드서베이'가 선거에 관한 여론조사를 공표하거나 보도 목적으로 실시하려면 중앙선거관리위원회규칙에 따른 요건을 충족하고 관할 선거여론조사심의위원회에 등록해야 하며, 등록증이 교부된 이후 기관에 관한 정보는 일반 국민에게 공개된다.\\ \vspace{0.5em}
\textbf{Self-Ask} \\ 
Are follow up questions needed here: Yes. (Continue) \\ \vspace{0.5em}
\textbf{IRCoT} \\
'트렌드서베이'는 \textcolor{red}{조사 시스템과 분석 전문 인력을 갖추고, 중앙선거관리위원회규칙에서 정하는 요건을 충족한 후 관할 선거여론조사심의위원회에 서면으로 등록을 신청해야 하며}, 관할 선거여론조사심의위원회는 신청을 \textcolor{blue}{접수한 날부터 7일 이내에 등록을 수리하고 등록증을 교부해야 한다.}\\
\bottomrule
\end{tabular}
\caption{Case study of a single Korean instance from \datasetName{}. The contents of each gold supporting provision are highlighted in different colors. While both SP and CoT capture the overall reasoning flow, they fail to identify the specific contents of the provisions. Self-Ask fail to answer the question. In IRCoT, one of the gold provisions is missing.}
\label{fig:casestudy3}
\end{figure*}

\begin{figure*}[ht]
\small
\begin{tabular}{p{\linewidth}}  
\toprule
\textbf{Background Scenario} \\ \midrule
갑은 신생 여론조사기관 '트렌드서베이'의 대표로, 다가오는 지방선거와 관련된 여론조사를 실시한 후 그 결과를 언론에 공표하려는 계획을 가지고 있다. 이를 위해 갑은 조사 시스템과 분석 전문 인력을 갖추고, 중앙선거관리위원회규칙에서 요구하는 요건을 충족한 뒤 관할 선거여론조사심의위원회에 등록을 서면으로 신청하였다. 등록 절차가 완료되면 해당 기관에 관한 정보가 일반 국민에게도 공개된다는 점을 인지하고 있다.\\ \midrule
\textbf{Question} \\ \midrule
'트렌드서베이'가 선거에 관한 여론조사를 공표하거나 보도 목적으로 실시하려면 어떤 요건과 절차를 거쳐야 하며, 갑의 등록 신청 이후 관할 선거여론조사심의위원회는 어떤 처리 절차를 따라야 하는가? 또한, 등록증이 교부된 이후 '트렌드서베이'에 관한 정보는 어떻게 처리되는가?\\ \midrule
\textbf{Gold Answer} \\ \midrule
'트렌드서베이'는 \textcolor{red}{조사 시스템, 분석 전문 인력 등 중앙선거관리위원회규칙으로 정한 요건을 갖추고 관할 선거여론조사심의위원회에 서면으로 등록을 신청해야 한다.} 관할 선거여론조사심의위원회는 \textcolor{blue}{등록신청을 접수한 날부터 7일 이내에 등록을 수리하고 등록증을 교부해야 한다.} 등록증이 교부되면, 선거여론조사심의위원회는 해당 기관에 관한 정보를 중앙선거관리위원회규칙으로 정하는 바에 따라 \textcolor{orange}{지체 없이 중앙선거여론조사심의위원회 홈페이지에 공개해야 한다.}\\ \midrule
\textbf{Gold supporting provisions} \\ \midrule
\textcolor{red}{공직선거법 8조 9 여론조사 기관ㆍ단체의 등록 등 1항} \\
① 여론조사 기관ㆍ단체가 공표 또는 보도를 목적으로 선거에 관한 여론조사를 실시하려는 때에는 조사시스템, 분석전문인력, 그 밖에 중앙선거관리위원회규칙으로 정하는 요건을 갖추어 관할 선거여론조사심의위원회에 서면으로 그 등록을 신청하여야 한다.\\
\textcolor{blue}{공직선거법 8조 9 여론조사 기관ㆍ단체의 등록 등 2항}\\
② 제1항에 따른 등록신청을 받은 관할 선거여론조사심의위원회는 그 신청을 접수한 날부터 7일 이내에 등록을 수리하고 등록증을 교부하여야 한다.\\
\textcolor{orange}{공직선거법 8조 9 여론조사 기관ㆍ단체의 등록 등 3항} \\
③ 선거여론조사심의위원회는 제2항에 따라 등록증을 교부한 여론조사 기관ㆍ단체(이하 "선거여론조사기관"이라 한다)에 관한 정보로서 중앙선거관리위원회규칙으로 정하는 정보를 지체 없이 중앙선거여론조사심의위원회 홈페이지에 공개하여야 한다.
\\ \midrule
\textbf{\methodName{} (Ours)} \\
'트렌드서베이'가 선거에 관한 여론조사를 공표하거나 보도 목적으로 실시하려면 \textcolor{red}{조사 시스템, 분석 전문 인력, 중앙선거관리위원회규칙으로 정하는 요건을 갖추어 관할 선거여론조사심의위원회에 서면으로 등록을 신청해야 한다.} 갑의 등록 신청 이후, 관할 선거여론조사심의위원회는 \textcolor{blue}{신청을 접수한 날부터 7일 이내에 등록을 수리하고 등록증을 교부해야 한다.} 등록증이 교부된 이후, '트렌드서베이'에 관한 정보는 중앙선거관리위원회규칙으로 정하는 정보를 \textcolor{orange}{지체 없이 중앙선거여론조사심의위원회 홈페이지에 공개해야 한다.}\\ \vspace{0.5em}
\textbf{FLARE} \\
'트렌드서베이'는 \textcolor{red}{조사 시스템과 분석 전문 인력을 갖추고, 중앙선거관리위원회규칙에서 정하는 요건을 충족한 후 관할 선거여론조사심의위원회에 서면으로 등록을 신청해야 하며}, 등록 신청 후 관할 위원회는 요건 충족 여부를 검토하여 등록증을 교부한다. \\ \vspace{0.5em}
\textbf{ProbTree} \\
'트렌드서베이'가 선거에 관한 여론조사를 공표하거나 보도 목적으로 실시하려면 \textcolor{red}{조사시스템, 분석전문인력 등 중앙선거관리위원회규칙으로 정한 요건을 갖추고 관할 선거여론조사심의위원회에 서면으로 등록을 신청해야 한다.} 등록 신청 후, 관할 선거여론조사심의위원회는 관련 절차에 따라 등록 여부를 심사하고, 등록증이 교부된 이후에는 '트렌드서베이'에 관한 정보가 관리된다.\\ \vspace{0.5em}
\textbf{BeamAggr} \\ 
선거에 관한 여론조사를 공표하거나 보도 목적으로 실시하려면, 여론조사 기관은 중앙선거관리위원회규칙으로 정한 요건을 갖추고 관할 선거여론조사심의위원회에 서면으로 등록을 신청해야 하며, 휴대전화 가상번호를 사용할 수 있습니다. \\ 
\bottomrule
\end{tabular}
\caption{Case study of a single Korean instance from \datasetName{}. The contents of each gold supporting provision are highlighted in different colors. FLARE and ProbTree correctly refer to one provision but miss the other two. BeamAggr provides a vague explanation and includes a hallucinated answer.}
\label{fig:casestudy4}
\end{figure*}
\section{Additional Results}
\label{smallresults}
\begin{table*}[ht]
\centering
\resizebox{\linewidth}{!}{%
\begin{tabular}{@{}ccccccccc@{}}
\toprule
            & \multicolumn{2}{c}{\cellcolor[HTML]{DAE8FC}F-1} & \multicolumn{2}{c}{\cellcolor[HTML]{DAE8FC}EM} & \multicolumn{2}{c}{\cellcolor[HTML]{FFFFC7}Token F-1} & \multicolumn{2}{c}{\cellcolor[HTML]{FFFFC7}\evalName{}} \\ \midrule
            & Qwen-8B  & EXAONE-7.8B  & Qwen-8B  & EXAONE-7.8B   & Qwen-8B   & EXAONE-7.8B   & Qwen-8B   & EXAONE-7.8B    \\
\small SP$^{\diamondsuit}$ \cite{NEURIPS2020_1457c0d6}   & -    & -    & -    & -   & 25.59   & 15.98   & 35.13    & 44.82     \\
\small CoT$^{\diamondsuit}$ \cite{NEURIPS2022_9d560961}  & -    & -    & -    & -   & 22.89   & 18.94   &  31.28   &   38.58   \\
\small SP \cite{NEURIPS2020_1457c0d6} + OR$^{\heartsuit}$  & 21.50  & 21.50    & 7.08   & 7.08    & 27.32   & 16.79   &  37.64   & 43.20     \\
\small CoT \cite{NEURIPS2022_9d560961} + OR$^{\heartsuit}$ & 21.50  & 21.50    & 7.08   & 7.08    & 26.68   & 22.64   &  39.43   &  43.57    \\
\small Self-Ask$^{\spadesuit}$ \cite{press-etal-2023-measuring} &  10.03  &  4.57  &  2.21  & 0.88  & 6.19  & 9.12    & 17.63 &  26.99    \\
\small IRCoT$^{\spadesuit}$ \cite{trivedi-etal-2023-interleaving} &  21.79  & 19.19   &  4.87  & 3.98  & 12.72  & 19.19    &  20.45   & 24.16     \\
\small FLARE$^{\spadesuit}$ \cite{jiang-etal-2023-active} & 25.40   & 20.38   & 5.31   &  3.10 &  22.72 & 11.84    & 38.84    &  23.35    \\
\small ProbTree$^{\clubsuit}$ \cite{cao-etal-2023-probabilistic}  & 13.26   &  7.76  & 3.10   & 0.88  & 28.53  &  19.61   & 40.93    &  42.30    \\
\small BeamAggr$^{\clubsuit}$ \cite{chu-etal-2024-beamaggr} & 11.27   & 5.94   & 2.65   & 0.88  & 17.75  &  10.90   &  33.35   & 31.60     \\ \midrule
\small \methodName$^{\clubsuit}$ (Ours)     & \textbf{34.18}   & \textbf{36.78}   &  \textbf{11.50}  &  \textbf{10.62} & \textbf{34.64}  & \textbf{24.40}    &  \textbf{49.41}   &  \textbf{52.39}    \\ \bottomrule
\end{tabular}%
}
\caption{Experimental results of various retrieval-augmented generation methods on \datasetName. Columns shaded in \colorbox[HTML]{ECF4FF}{blue} measure \textbf{retrieval accuracy}, and columns shaded in \colorbox[HTML]{FFFC9E}{yellow} measure \textbf{generation accuracy}. Best results are highlighted in \textbf{bold}. We utilize Qwen3-8B \cite{qwen3} and EXAONE-3.5-7.8B \cite{exaone-3.5}.
($\diamondsuit$: No-retrieval, $\heartsuit$: One-time retrieval, $\spadesuit$: Iterative retrieval, $\clubsuit$: Sub-query retrieval).}
\label{tab:small_result}
\end{table*}
Table \ref{tab:small_result} shows experimental results on \datasetName{} with smaller LLMs such as Qwen3-8B \cite{qwen3} and EXAONE-3.5-7.8B \cite{exaone-3.5}.
The results show consistent patterns with robust LLMS as shown in the main body of the paper represented in Table \ref{tab:main1}.
\methodName{} reliably achieves the highest performance across metrics.
These findings confirm that our approach is effective even with limited model capacity, highlighting its scalability and generalization.
\end{document}